\documentclass[11pt]{article}

\usepackage[preprint]{acl}

\usepackage{listings}
\usepackage{enumitem}
\usepackage{times}
\usepackage{latexsym}
\usepackage{booktabs}
\usepackage{pifont}
\usepackage{xcolor}
\usepackage{amssymb}
\usepackage{tabularx}
\usepackage{array}
\usepackage{makecell}
\usepackage{multirow}
\usepackage{amsmath}
\usepackage{siunitx}
\usepackage[T1]{fontenc}
\definecolor{ao(english)}{rgb}{0.0, 0.5, 0.0}

\newcommand{\cmark}{\textcolor{ao(english)}{\large\ding{51}}} 
\newcommand{\xmark}{\textcolor{red}{\large\ding{55}}}       

\usepackage[utf8]{inputenc}

\usepackage{microtype}

\usepackage{inconsolata}

\usepackage{graphicx}

\newcommand{\benchmark}{OmniACBench}

\title{\benchmark: A Benchmark for Evaluating Context-Grounded \\ Acoustic Control in Omni-Modal Models}

\author{
\textbf{Seunghee Kim}$^{1}$\thanks{This work was done during the residency program at NAVER Cloud.},
\textbf{Bumkyu Park}$^{2}$\footnotemark[1],
\textbf{Kyudan Jung}$^{3}$\footnotemark[1], 
\textbf{Joosung Lee}$^{4}$, \\
\textbf{Soyoon Kim}$^{4}$, 
\textbf{Jeonghoon Kim}$^{3,4}$,
\textbf{Taeuk Kim}$^{1}$\thanks{Co-corresponding authors.},
\textbf{Hwiyeol Jo}$^{4}$\footnotemark[2] \\
$^{1}$Hanyang University,
$^{2}$Seoul National University,
$^{3}$KAIST AI,
$^{4}$NAVER Cloud \\
\texttt{\{gyg9325, kimtaeuk\}@hanyang.ac.kr,}\\
\texttt{bumkyu00@europa.snu.ac.kr, kyudan@kaist.ac.kr,}\\
\texttt{\{rung.joo, soyoon.kim, jeonghoon.samuel, hwiyeol.jo\}@navercorp.com}
}

\begin{document}
\maketitle
\begin{abstract}
Most testbeds for omni-modal models assess multimodal understanding via textual outputs, leaving it unclear whether these models can properly \textit{speak} their answers.
To study this, we introduce \benchmark, a benchmark for evaluating \textit{context-grounded acoustic control} in omni-modal models.
Given a spoken instruction, a text script, and an image, a model must read the script aloud with an appropriate tone and manner.
\benchmark\ comprises 3,559 verified instances covering six acoustic features: speech rate, phonation, pronunciation, emotion, global accent, and timbre.
Extensive experiments on eight models reveal their limitations in the proposed setting, despite their strong performance on prior textual-output evaluations.
Our analyses show that the main bottleneck lies not in processing individual modalities, but in integrating multimodal context for faithful speech generation.
Moreover, we identify three common failure modes---weak direct control, failed implicit inference, and failed multimodal grounding---providing insights for developing models that can effectively verbalize responses.


\end{abstract}

\section{Introduction}
Multimodal Large Language Models (MLLMs) have rapidly evolved from bi-modal systems---such as text–vision \cite{alayrac2022flamingo, li2023blip} and text–audio \cite{deshmukh2023pengi, tang2023salmonn}---to omni-modal architectures that jointly process text, vision, and audio as input \cite{han2024onellm, li2024baichuan}.\footnote{\textbf{Omni-modal}: all of text, vision, and audio. \textbf{Multimodal}: any combination of these three modalities.}
More recently, this trend has extended beyond inputs: modern omni-modal models can now generate speech responses as well as text \cite{xu2025qwen25omnitechnicalreport, xu2025qwen3, wang2025mgm, tong2025interactiveomni}.
This shift marks a transition for omni-modal systems from multimodal understanding to generating responses in diverse modalities.

\begin{figure}[t]  
\centering
\includegraphics[width=1\columnwidth, keepaspectratio]{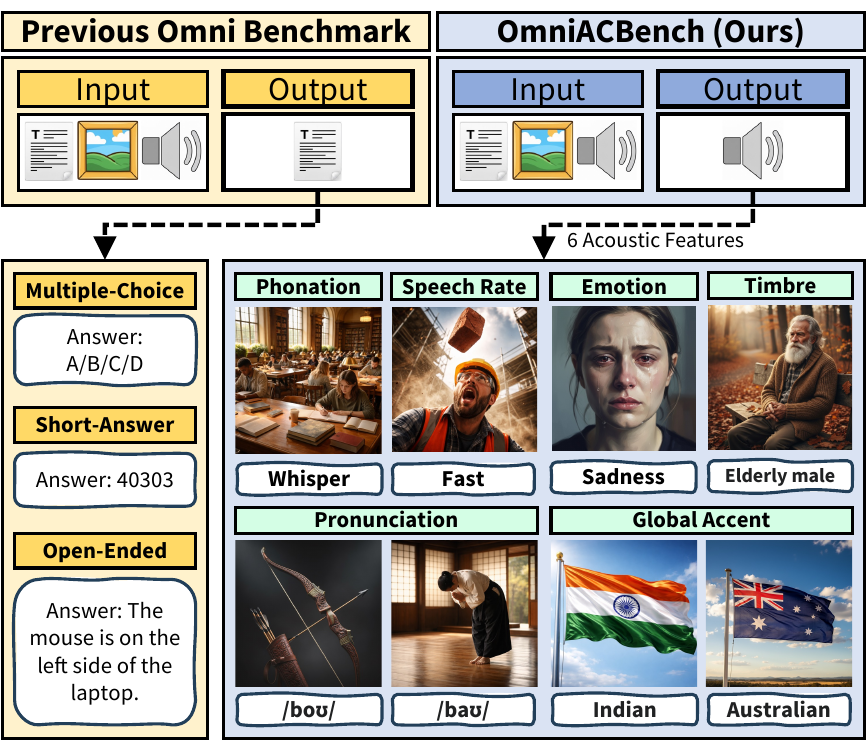}
  \caption{
Comparison of prior omni-modal benchmarks and \benchmark. Existing ones assess multimodal understanding via text outputs, whereas ours targets speech generation given text, vision, and speech inputs.
}
\label{fig:previous_bench_vs_ours}
\end{figure}

\begin{table}[t]
\centering
\small
\renewcommand{\arraystretch}{0.7}
\setlength{\tabcolsep}{2.7pt}
\begin{tabular}{l ccc c l}
\toprule
\multirow{2}{*}[-0.6ex]{\textbf{Benchmark}}
  & \multicolumn{3}{c}{\textbf{Input}}
  & \multirow{2}{*}[-0.6ex]{\textbf{Output}}
  & \multirow{2}{*}[-0.6ex]{\textbf{Target Features}} \\
\cmidrule(lr){2-4}
 & \textbf{T} & \textbf{V} & \textbf{S} & & \\
\midrule
OmniBench        & \cmark & \cmark & \cmark & Text & Tri-Modal \\
WorldSense       & \cmark & \cmark & \cmark & Text & Real World \\
Daily-Omni       & \cmark & \cmark & \cmark & Text & Temporal \\
OMHBench         & \cmark & \cmark & \cmark & Text & Multi Hop \\
OmniVideoBench   & \cmark & \cmark & \cmark & Text & Diverse Video \\
UNO-Bench        & \cmark & \cmark & \cmark & Text & Uni/Omni Link \\
AV-SpeakerBench  & \cmark & \cmark & \cmark & Text & Speaker Centric \\
FutureOmni      & \cmark & \cmark & \cmark & Text & Future Forecast \\
\midrule
URO-Bench        & \xmark & \xmark & \cmark & Speech & Em, Si, Re \\
VocalBench       & \xmark & \xmark & \cmark & Speech & Em \\
\multirow{2}{*}{S2S-Arena}
 & \multirow{2}{*}{\xmark} & \multirow{2}{*}{\xmark} & \multirow{2}{*}{\cmark}
 & \multirow{2}{*}{Speech}
 & Em, Ti, SR \\
 & & & & & Pr, Si, St \\
ParaS2SBench     & \xmark & \xmark & \cmark & Speech & Em, Ti, Sa \\
VA-Eval-Viewing  & \xmark & \cmark  & \cmark & Speech & Visual QA \\
VA-Eval-Speaking & \xmark & \xmark & \cmark & Speech & Em, Ti \\
\midrule
\multirow{2}{*}{\textbf{\benchmark}}
 & \multirow{2}{*}{\cmark} & \multirow{2}{*}{\cmark} & \multirow{2}{*}{\cmark}
 & \multirow{2}{*}{Speech}
 & Em, Ti, SR, \\
 & & & & & Pr, GA, Ph \\
\bottomrule
\end{tabular}
\caption{
Comparing omni-modal and speech generation benchmarks by input/output modalities and target features.
\textbf{Abbreviations:} \textbf{T}ext/\textbf{V}ision/\textbf{S}peech, \textbf{Em}otion, \textbf{Ti}mbre, \textbf{S}peech \textbf{R}ate, \textbf{Pr}onunciation, \textbf{G}lobal \textbf{A}ccent, \textbf{Ph}onation, \textbf{Si}nging, \textbf{Re}citation, \textbf{St}ress, \textbf{Sa}rcasm.
}
\label{tab:related_work_benchmarks_compare}
\end{table}

Alongside these advances, several benchmarks have been introduced to evaluate omni-modal models \cite{li2024omnibench, hong2025worldsense, zhou2025daily, kim2025omhbench, chen2026futureomni}.
Most of them focus on testing multimodal understanding, examining whether systems can interpret multimodal inputs and produce semantically correct textual answers.
As a result, they provide valuable testbeds for measuring models’ comprehension and reasoning abilities across modalities.
However, as omni-modal models increasingly generate outputs in multiple formats, a key research question remains: \textbf{what should be considered and evaluated when responses are delivered in speech?}


Speech responses encode information not only in words but also in acoustic delivery, such as speaking rate, phonation, and other paralinguistic cues \cite{guyer2021paralinguistic}.
Therefore, speech-based evaluation should consider both whether the content is correct and whether the delivery is appropriate---a dimension with no direct counterpart in text-based evaluation.
The challenge becomes especially pronounced in omni-modal settings, where appropriate acoustic realization must be inferred from signals across different modalities. 
For example, a conversation in a library may call for whisper-like phonation, an emergency situation for a faster speaking rate, and a sorrowful scene for a sad vocal expression.
We refer to the ability to generate acoustically appropriate speech given multimodal context as \textbf{context-grounded acoustic control}.

To test this aspect, we propose \textbf{\benchmark}, a benchmark for context-grounded \textbf{a}coustic \textbf{c}ontrol in \textbf{omni}-modal models.\footnote{\benchmark\ is available at \url{https://huggingface.co/datasets/SeungHeeKim/OmniACBench}.} 
Given a spoken instruction, a text script, and an image, the model must generate speech that faithfully reads the script while realizing an acoustic delivery consistent with the combined multimodal context.
\benchmark\ covers six acoustic features and evaluates both measurable and abstract properties of speech output. 
Figure~\ref{fig:previous_bench_vs_ours} highlights the key differences between prior omni-modal datasets and \benchmark.

We evaluate eight models on \benchmark\ and find that current systems perform poorly overall, even when they achieve strong results on text-based evaluations.
Our analysis further reveals that this gap does not stem simply from failures in processing a specific modality, but from the difficulty of generating speech that preserves the target content while acoustically reflecting multimodal context.


\begin{figure*}[t]  
\centering
\includegraphics[width=1\textwidth, keepaspectratio]{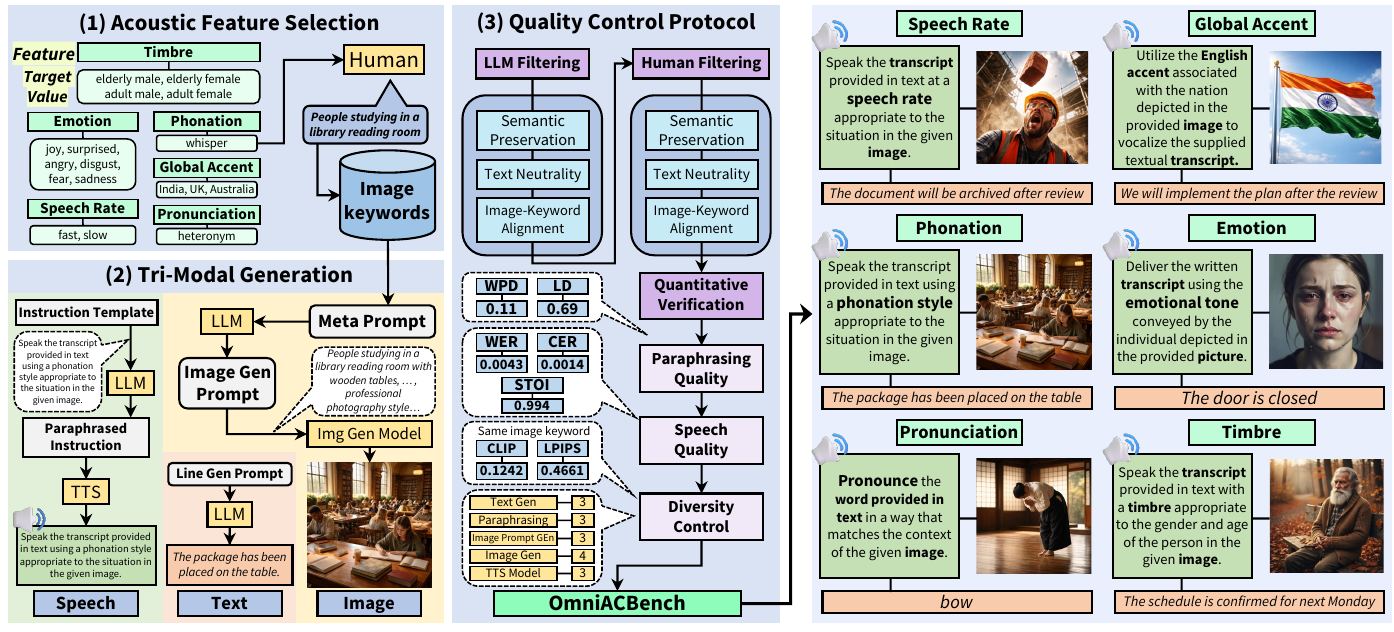}
  \caption{
Construction pipeline of \benchmark\ with representative examples for each acoustic feature.
(1) \textbf{Acoustic Feature Selection} defines the target acoustic features and associated image keywords.
(2) \textbf{Tri-Modal Generation} constructs each instance from a neutral text script, a spoken control signal, and a generated image.
(3) \textbf{Quality Control Protocol} applies filtering and quantitative verification to ensure data quality and diversity.
}
\label{fig:dataset_construction_pipeline}
\end{figure*}

\section{Related Work}

\subsection{Omni-Modal Benchmarks}
Several benchmarks have been introduced to assess omni-modal capabilities (see Table~\ref{tab:related_work_benchmarks_compare}). 
OmniBench \cite{li2024omnibench}, WorldSense \cite{hong2025worldsense}, Daily-Omni \cite{zhou2025daily}, OmniVideoBench \cite{li2025omnivideobench}, and AV-SpeakerBench \cite{nguyen2025see} examine how well systems understand inputs spanning text, vision, and audio. 
OMHBench \cite{kim2025omhbench} further addresses modality shortcut issues by introducing omni-modal multi-hop reasoning, while UNO-Bench \cite{chen2025uno} shows that omni-modal capability is jointly determined by underlying uni-modal abilities. FutureOmni \cite{chen2026futureomni} focuses on forecasting future events grounded in audio-visual context. 
Most existing benchmarks study multimodal understanding through text outputs, whereas \benchmark\ shifts the focus to speech-based evaluation, assessing both semantic fidelity and context-grounded acoustic control.

\subsection{Speech Generation Benchmarks}
Recent benchmarks for speech generation have moved beyond textual correctness to evaluate how spoken responses are delivered. 
URO-Bench \cite{yan2025uro} and VocalBench \cite{liu2025vocalbench} assess broad speech-interaction abilities, including acoustic and paralinguistic phenomena. 
S2S-Arena \cite{jiang2025s2s} and ParaS2SBench \cite{yang2025paras2s} focus more directly on speech-to-speech instruction following, examining whether models produce appropriate content and speaking style under spoken cues. 
VoiceAssistant-Eval \cite{wang2025voiceassistant} further expands this line to voice-assistant scenarios spanning listening, speaking, and viewing. 
However, these benchmarks primarily evaluate speech-centric interaction or general spoken assistant behavior, whereas \benchmark\ focuses on context-grounded acoustic control in speech generation under joint text, vision, and speech inputs.

\section{\benchmark}

We present \textbf{\benchmark}, a benchmark for evaluating context-grounded acoustic control in omni-modal models.
Each instance in the dataset comprises a text script, a spoken instruction, and an image, requiring the model to read the script aloud with appropriate acoustic realization.
The three input modalities play distinct roles: the \textit{text} defines the linguistic content, the \textit{spoken instruction} specifies the acoustic aspect to control, and the \textit{image} provides contextual cues.
By separating content, control, and contextual grounding across modalities, \benchmark\ evaluates \textit{context-grounded acoustic control} rather than treating the task as prosodic rendering or explicit style conditioning.

The construction of \benchmark\ follows a three-stage pipeline: (1) \textbf{Acoustic Feature Selection}, (2) \textbf{Tri-Modal Generation}, and (3) \textbf{Quality Control Protocol}, as illustrated in Figure \ref{fig:dataset_construction_pipeline}.

\subsection{Acoustic Feature Selection}
In the first stage, we select acoustic features and define their target values, each representing a specific realization (e.g., \textit{fast} for Speech Rate or \textit{angry} for Emotion). 
We choose these features based on two complementary criteria.

First, we consider \textbf{multimodal groundability}: each feature should admit natural visual grounding, so that images can provide meaningful clues for the intended target value (e.g., an emergency scene implicitly calling for a faster speech rate).
Second, we consider \textbf{evaluation diversity}: the selected features should span different forms of acoustic evaluation.
Some features correspond to measurable attributes that can be assessed with explicit objective metrics (e.g., Speech Rate measured in words per minute), whereas others are abstract and cannot be reduced to a single objective metric (e.g., Emotion).
Including both types enables a more comprehensive evaluation of acoustic control.

Based on these criteria, we select six acoustic features: \textbf{Speech Rate}, \textbf{Phonation}, \textbf{Pronunciation}, \textbf{Emotion}, \textbf{Global Accent}, and \textbf{Timbre}.\footnote{Refer to Appendix~\ref{app:additional_acoustic_features} for additional feature candidates.}
Each feature is associated with a predefined set of target values, and each benchmark instance is defined by a \textbf{feature--value} pair (e.g., Emotion--angry, Speech Rate--fast).
For each target value, human annotators manually curate a set of image keywords describing visual scenes or concepts naturally associated with it, such as ``people studying in a library reading room'' for \textit{Phonation--whisper} and ``a boiling-over pot'' for \textit{Speech Rate--fast}.

\begin{table}[h]
  \centering
  \small
  \begin{tabular}{lr}
    \toprule
    \textbf{Category} & \textbf{Count} \\
    \midrule
    Total Instances & 3,559 \\
    \midrule
    \textbf{Emotion} & 597 \\
    \hspace{1em}Anger & 100 \\
    \hspace{1em}Disgust & 98 \\
    \hspace{1em}Fear & 100 \\
    \hspace{1em}Joy & 99 \\
    \hspace{1em}Sadness & 100 \\
    \hspace{1em}Surprise & 100 \\
    \midrule
    \textbf{Global Accent (Region)} & 549 \\
    \hspace{1em}Australia & 200 \\
    \hspace{1em}India & 198 \\
    \hspace{1em}UK & 151 \\
    \midrule
    \textbf{Speech Rate} & 596 \\
    \hspace{1em}Fast & 298 \\
    \hspace{1em}Slow & 298 \\
    \midrule
    \textbf{Timbre} & 596 \\
    \hspace{1em}Adult Female & 147 \\
    \hspace{1em}Adult Male & 150 \\
    \hspace{1em}Elderly Female & 150 \\
    \hspace{1em}Elderly Male & 149 \\
    \midrule
    \textbf{Phonation} & 590 \\
    \hspace{1em}Whisper & 590 \\
    \midrule
    \textbf{Pronunciation} & 631 \\
    \hspace{1em}Heteronym & 631 \\
    \midrule
    Avg. Image Resolution & 512$\times$512 \\
    Avg. Speech Duration (s) & 6.8 \\
    Avg. Text Length (chars) & 36.8 \\
    \bottomrule
  \end{tabular}
  \caption{Detailed statistics of \benchmark.}
  \label{tab:dataset_statistics}
\end{table}

\subsection{Tri-Modal Generation}
In the second phase, we construct the three modalities of each \benchmark\ instance---\textit{text}, \textit{speech}, and \textit{image}---from a given feature--value pair.
All instances are synthetically generated using generative models, enabling scalable and controllable construction.
This design also supports fine-grained analysis of model behavior (Sections~\ref{subsec:basic-cap-check} and~\ref{subsec:task-decomp}).

\paragraph{Text}
The text modality provides the script that the model is required to read aloud.
To prevent shortcut learning, the script is designed to remain neutral and not directly encode the intended acoustic characteristic.
We thus generate scripts using an LLM, explicitly prompting it to produce content that is neutral with respect to attributes such as emotion, nationality, gender, and age.


\begin{table}[t]
\centering
\small
\setlength{\tabcolsep}{3pt}
\begin{tabular}{@{}lll@{}}
\toprule
\textbf{Aspect} & \textbf{Metric} & \textbf{Value} \\
\midrule
Paraphrase & (MRPC) WPD$\uparrow$ / LD$\uparrow$ & 0.12 / 0.42 \\
 & (PAWS) WPD$\uparrow$ / LD$\uparrow$ & 0.07 / 0.13 \\
 & (Ours) WPD$\uparrow$ / LD$\uparrow$ & 0.11 / 0.69 \\
\midrule
Speech & WER$\downarrow$ / CER$\downarrow$ & 0.004 / 0.001 \\
 & STOI$\uparrow$ & 0.994 \\
\midrule
Image & (keyword) CLIP$\uparrow$ / LPIPS$\uparrow$ & 0.067 / 0.373 \\
 & (Ours) CLIP$\uparrow$ / LPIPS$\uparrow$ & 0.124 / 0.466 \\
\bottomrule
\end{tabular}
\caption{
Quantitative verification of dataset construction quality.
Higher WPD and LD denote greater paraphrase diversity; lower WER and CER and higher STOI indicate better speech quality; higher CLIP distance and LPIPS reflect greater image diversity.
}
\label{tab:quantitative_verification}
\end{table}

\paragraph{Speech}
The speech modality serves as a control signal that specifies which acoustic dimension should govern delivery.
The model must use this signal to determine which acoustic property to infer from the image and realize in speech while reading the text.
For each acoustic feature, we first design an instruction template and then use an LLM to generate instance-level paraphrases that increase linguistic diversity while preserving semantic intent.
The resulting paraphrases are then synthesized into speech using a text-to-speech (TTS) model.

\paragraph{Image}
The image modality provides situational cues for inferring the target value within the acoustic dimension specified by the spoken instruction.
Directly using the image keywords from the previous stage as prompts would limit both diversity and scalability.
To address this, we design a meta-prompt that instructs an LLM to expand each keyword into an image-generation prompt in which the keyword remains the central visual focus while additional scene attributes are introduced.
Specifically, the LLM is guided to produce a prompt consisting of 5 to 8 comma-separated visual elements.
These prompts are then used by image generation models to synthesize the final images.

\begin{table*}[t]
\centering
\small
\setlength{\tabcolsep}{4pt}
\begin{tabular}{@{}p{0.23\textwidth}p{0.65\textwidth}@{}}
\toprule
\textbf{Task} & \textbf{Models} \\
\midrule
Text Generation &
Gemini 3 Flash, GPT 5 Nano, Claude Haiku 4.5 \\

Transcript Paraphrasing &
Gemini 3 Flash, GPT 5 Nano, Claude Haiku 4.5 \\

Image Prompt Generation &
Gemini 3 Flash, GPT 5 Nano, Claude Haiku 4.5 \\

Image Generation &
Gemini 3 Pro Image, Gemini 2.5 Flash Image, GPT Image 1, GPT Image 1.5 \\

Text-to-Speech &
Eleven Multilingual v2, Eleven Flash v2.5, Eleven Turbo v2.5 \\

LLM-Based Filtering &
Grok 4 Fast \\
\bottomrule
\end{tabular}
\caption{
Model pools employed at each stage of dataset construction. For every generation call, a single model is sampled uniformly at random from the pool of the corresponding stage.
}
\label{tab:model_pools}
\end{table*}

\subsection{Quality Control Protocol}
\label{subsec:quality-control}
We apply a quality control protocol with two components: \textbf{Data Filtering} and \textbf{Quantitative Verification}.
These procedures aim to detect and remove potential negative artifacts that may arise during the synthetic generation of \benchmark.\footnote{Refer to Appendix \ref{app:detail-qcp} for the full details of our protocol.}

\subsubsection{Data Filtering}
We use three filtering strategies (Semantic Preservation, Text Neutrality, and Image–Keyword Alignment) applied at different stages of the pipeline.
We first perform LLM-based filtering, followed by human verification using the same methods to double-check the results.
\textbf{Semantic Preservation} tests whether paraphrased spoken instructions preserve the intent of the original sentence.
\textbf{Text Neutrality} removes scripts whose content alone reveals the target acoustic value, which could otherwise enable shortcut learning.
\textbf{Image--Keyword Alignment} checks whether generated images remain semantically consistent with their original keywords.

Starting from 3,640 generated instances, 3,586 remain after LLM-based filtering and 3,559 after human verification, yielding a final retention rate of 97.78\%.
The benchmark spans six acoustic features with roughly 600 instances each, indicating a balanced feature distribution.
Detailed statistics are summarized in Table~\ref{tab:dataset_statistics}.
Figure~\ref{fig:app-data-examples} in the Appendix presents concrete examples from \benchmark.

\subsubsection{Quantitative Verification}
\label{subsubsec:quantitative-verification}
To further substantiate the quality of the benchmark, we conduct a quantitative verification, the results of which are summarized in Table~\ref{tab:quantitative_verification}.

\paragraph{Paraphrasing Quality}
We evaluate the linguistic diversity of paraphrased spoken instructions using Word Position Deviation (WPD) and Lexical Deviation (LD) \cite{liu2022towards}.
\benchmark\ achieves WPD/LD scores of 0.11/0.69, compared to 0.12/0.42 on MRPC \cite{dolan2005automatically} and 0.07/0.13 on PAWS \cite{zhang2019paws}, two standard paraphrasing datasets.
These results suggest sufficient linguistic diversity in the benchmark.

\paragraph{Speech Quality}
We evaluate synthesized speech using WER ($\downarrow$) and CER($\downarrow$) computed by Whisper-large-v3 \cite{radford2023robust} to measure transcription fidelity, and STOI ($\uparrow$) to gauge intelligibility \cite{kumar2023torchaudio}.
The resulting scores (WER 0.004, CER 0.001, and STOI 0.994) indicate near-perfect transcription fidelity and intelligibility.

\begin{table*}[t]
\centering
\small
\setlength{\tabcolsep}{3.3pt}
\begin{tabular}{lccccccc}
\toprule

&
\multicolumn{1}{c}{\textbf{Semantic}} &
\multicolumn{1}{c}{\textbf{Speech Rate}} &
\multicolumn{1}{c}{\textbf{Pron.}} &
\multicolumn{1}{c}{\textbf{Phonation}} &
\multicolumn{1}{c}{\textbf{Emotion}} &
\multicolumn{1}{c}{\textbf{Global Accent}} &
\multicolumn{1}{c}{\textbf{Timbre}} \\

\cmidrule(lr){2-2}
\cmidrule(lr){3-3}
\cmidrule(lr){4-4}
\cmidrule(lr){5-5}
\cmidrule(lr){6-6}
\cmidrule(lr){7-7}
\cmidrule(lr){8-8}

\textbf{Models} & WER $\downarrow$ & $\Delta$WPM $\uparrow$ & PER $\downarrow$ & VFR@0.3 $\uparrow$ (\%) & Emo-Acc $\uparrow$ (\%) & GA-Acc $\uparrow$ (\%) & Tim-Acc $\uparrow$ (\%) \\
\midrule

\textit{Reference Scores}
& 0.05 & 65.87 & 1.21 & 96.78 & 89.43 & 97.29 & 96.67 \\
\midrule

MiniCPM-o 4.5
& \textbf{1.04} & \textbf{6.42} & \textbf{5.46} & \textbf{1.69} & \textbf{21.44} & \textbf{39.34} & 24.66 \\

InteractiveOmni 8B
& \underline{1.23} & -0.73 & \underline{6.46} & 0.00 & 14.57 & 33.70 & 24.33 \\

InteractiveOmni 4B
& 1.31 & 0.45 & 7.97 & 0.00 & 15.81 & 33.33  & 24.83 \\

Qwen3-Omni 30B
& 2.14 & -1.81 & 7.40 & 0.00 & 17.09 & 31.33 & 25.17 \\

Qwen2.5-Omni 7B
& 4.15 & 0.76 & 10.27 & 0.85 & \underline{19.10} & 28.96 & 24.66 \\

Qwen2.5-Omni 3B
& 4.51 & \underline{0.91} & 13.47 & 0.00 & 16.58 & 28.71 & 24.66 \\

Uni-MoE-2.0-Omni
& 5.21 & -2.82 & 9.27 & \underline{1.69} & 16.75 & 36.25 & \underline{25.17} \\

MGM-Omni 7B
& 5.96 & -0.60 & 21.97 & 0.00 & 16.42 & \underline{36.61} & \textbf{25.34} \\

\bottomrule
\end{tabular}

\caption{
Main results on \benchmark. Semantic denotes semantic fidelity, and Pron. denotes pronunciation. WER measures semantic fidelity; $\Delta$WPM, PER, and VFR@0.3 evaluate speech rate, pronunciation, and phonation; Emo-Acc, GA-Acc, and Tim-Acc assess emotion, global accent, and timbre. The \textit{Reference Scores} denote those from human or trained evaluators. Random baseline performance for the last three features is 16.7\%, 33.3\%, and 25.0\%, respectively. Best in \textbf{bold}, second-best \underline{underlined}.
}
\label{tab:main_table}
\end{table*}

\paragraph{Diversity Control}

First, we use meta-prompting rather than na\"ive keyword prompting. 
Its effectiveness is verified by measuring the average pairwise CLIP embedding cosine distance \cite{radford2021learning} and LPIPS \cite{zhang2018unreasonable} within each keyword group. Meta-prompting yields higher values on both metrics (CLIP 0.1242 vs. 0.0671; LPIPS 0.4661 vs. 0.3733), indicating greater visual variation than keyword-only prompting.

Second, to reduce model-specific bias, we maintain diverse model pools for each generation stage---three for text generation, three for transcript paraphrasing, three for image prompt generation, four for image generation, and three for TTS.
For every generation call within the pipeline, we randomly sample one model from the corresponding pool.
LLM-based filtering is performed using a separate model.
The models used are listed in Table \ref{tab:model_pools}.


\section{Experiments}
\subsection{Evaluation Methods and Metrics}
\label{subsec:eval-method}

\paragraph{Semantic Fidelity}
We assess whether the generated speech preserves the target text.
To do so, we transcribe the generated speech using Whisper-large-v3 \cite{radford2023robust} and compute Word Error Rate (WER) against the reference script.

\paragraph{Measurable Acoustic Features}
\textbf{Speech Rate} is measured in Words Per Minute (WPM).
To account for differences in default speaking speed across models, we report $\Delta$WPM, defined as the difference between the average WPM of instances assigned the target values \textit{fast} and \textit{slow} (i.e., $\Delta\text{WPM} = \overline{\text{WPM}}_\text{fast} - \overline{\text{WPM}}_\text{slow}$).
A larger $\Delta$WPM indicates a clearer distinction between the two target values.
\textbf{Pronunciation} is evaluated using Phoneme Error Rate (PER). 
We extract phoneme sequences from generated speech using POWSM \cite{li2025powsm} and compare them with the reference phoneme sequence.
\textbf{Phonation} is evaluated via whisper detection.
Because whispered speech exhibits limited vocal-fold vibration and thus little or no fundamental frequency ($F_0$) \cite{zhao2016study,gudepu2020whisper}, we compute the Voiced Frame Ratio (VFR), i.e., the proportion of frames with detected $F_0$, and classify an utterance as whisper-like if VFR~$\leq$~0.3.
We validate this threshold on the Expresso dataset \cite{nguyen2023expresso}, where 90.8\% of whisper-style and only 0.2\% of normal-style utterances satisfy the criterion.
We report the resulting detection rate as VFR@0.3.

\paragraph{Abstract Acoustic Features}
Following prior work that evaluates using model-based evaluators \cite{yan2025uro, liu2025vocalbench, wang2025voiceassistant, yang2025emovoice}, we assess three features---\textbf{Emotion}, \textbf{Global Accent}, and \textbf{Timbre}---using task-specific evaluators trained for our label space.
For each feature, we collect speech samples from diverse sources (11k for Emotion, 12k for Global Accent, and 12k for Timbre) and train a WavLM-Large-based classifier \cite{chen2022wavlm}; the details are provided in Appendix~\ref{app:appendix-evaluator-train-detail}.
On held-out test sets, the evaluators achieve accuracies of 89.43\%, 97.29\%, and 96.67\% for Emotion, Global Accent, and Timbre, respectively.
We use these evaluators to score model-generated speech, reported as Emo-Acc, GA-Acc, and Tim-Acc.

\subsection{Experimental Setup}\label{sec:exp_setup}
We evaluate eight omni-modal models---MiniCPM-o 4.5 \citep{openbmb2026minicpmo}, InteractiveOmni (8B and 4B) \citep{tong2025interactiveomni}, Qwen3-Omni 30B \citep{xu2025qwen3}, Qwen2.5-Omni (7B and 3B) \citep{xu2025qwen25omnitechnicalreport}, Uni-MoE-2.0-Omni \citep{li2025unimoe20}, and MGM-Omni 7B \citep{wang2025mgm}---that take text, image, and speech inputs and generate speech outputs. 
Table \ref{tab:main_table} shows their performance on \benchmark.
We also report the \textbf{Reference Scores} for better interpretation: semantic fidelity and measurable features are computed from five human annotators, while abstract features are measured by held-out evaluator accuracy.

\subsection{Main Results}

\paragraph{Overall Trends}
The performance remains limited across models, highlighting the difficulty of \benchmark\ for current omni-modal models. 
Even models such as Qwen3-Omni 30B, which perform strongly on prior omni-modal benchmarks \cite{kim2025omhbench, li2025omnivideobench, chen2025uno, nguyen2025see, chen2026futureomni}, show weak performance on \benchmark. 
MiniCPM-o 4.5 achieves relatively stronger results on most metrics, yet remains far below the Reference Scores.
These results suggest that \benchmark\ captures capabilities that have not been sufficiently examined in existing benchmarks.

\paragraph{Semantic Fidelity}
We first examine whether models faithfully reproduce the target script during speech generation.
Compared with the Reference Scores, all models exhibit substantially higher WER, indicating that semantic fidelity itself becomes challenging when speech generation is conditioned on multimodal inputs.
That is, even before acoustic control is considered, models often fail to preserve the target text.

\paragraph{Measurable Acoustic Features}
Performance on measurable acoustic features---Speech Rate, Pronunciation, and Phonation---remains suboptimal for nearly all models. 
Most $\Delta$WPM values stay close to zero, suggesting little ability to modulate speaking rate even when the context implies a clear difference in tempo. 
Likewise, high PER scores and near-zero VFR@0.3 values indicate that models rarely realize fine-grained pronunciation control or whisper-like phonation. 
Overall, these results show that mapping contextual cues to precise, objectively measurable acoustic variation remains a major challenge for current omni-modal models.

\paragraph{Abstract Acoustic Features}
Compared with measurable features, abstract features appear somewhat easier to model, though overall performance remains limited.
MiniCPM-o 4.5 achieves clear gains over the random baselines on Emotion and Global Accent, suggesting that it can partially reflect contextually implied speaker attributes in generated speech.
However, most other models remain close to chance on these features, and timbre control is weak for all models.
In summary, current omni-modal models still struggle to ground contextual information into controllable acoustic attributes, even for high-level properties.

\section{Analysis}
\label{sec:analysis}
In this section, we investigate why current omni-modal models perform poorly on \benchmark. 
We first explore whether this difficulty arises from insufficient processing of specific modalities.
We then perform controlled input decomposition to identify failure points and characterize the resulting failure modes.
Finally, we conduct an additional linear probing analysis to examine whether multimodal context remains decodable up to the speech generation stage.

\subsection{Fundamental Capability Assessment}
\label{subsec:basic-cap-check}
To interpret whether poor performance on \benchmark\ reflects limitations in fundamental, modality-specific skills, we conduct three diagnostic ablation studies isolating core components: (1) script-only speech generation, (2) spoken-instruction conditioning, and (3) visual acoustic cue inference.
Table \ref{tab:basic_capability_v2} reports the results.

\begin{table}[t]
\centering
\footnotesize
\setlength{\tabcolsep}{1.5pt}
\begin{tabular}{l@{\hskip 1pt}ccc}
\toprule
\multirow{2}{*}{}
& \makecell{\textbf{Script-Onl}y} 
& \makecell{\textbf{Spoken}\\\textbf{Conditioning}} 
& \makecell{\textbf{Visual Cue}\\\textbf{Inference}} \\
\cmidrule(lr){2-2}
\cmidrule(lr){3-3}
\cmidrule(lr){4-4}
\textbf{Models} & WER $\downarrow$ & WER $\downarrow$ & Acc $\uparrow$ \\
\midrule
MiniCPM-o 4.5      & 0.12 & 0.12 & 94.75 \\
InteractiveOmni 8B & 0.12 & 0.13 & 95.56 \\
InteractiveOmni 4B & 0.10 & 0.11 & 92.88 \\
Qwen3-Omni 30B     & 0.11 & 0.13 & 97.50 \\
Qwen2.5-Omni 7B    & 0.13 & 0.13 & 94.21 \\
Qwen2.5-Omni 3B    & 0.14 & 0.15 & 92.19 \\
Uni-MoE-2.0-Omni   & 0.12 & 0.12 & 92.59 \\
MGM-Omni 7B        & 0.09 & 0.09 & 92.11 \\
\bottomrule
\end{tabular}
\caption{
Diagnostic results for three component capabilities: script-only speech generation, spoken-instruction conditioning, and visual acoustic cue inference.
Strong performance across these controlled settings suggests that weak performance on \benchmark\ is not explained by missing component abilities alone.
}
\label{tab:basic_capability_v2}
\end{table}

\paragraph{Script-Only Speech Generation}
We first test whether models can reliably read a provided script without multimodal grounding or acoustic control.
In this setting, models receive only the target script from \benchmark\ and generate speech.
WER is uniformly low across models, suggesting that literal script reading is not the primary bottleneck.

\paragraph{Spoken-Instruction Conditioning}
In this ablation, models receive a spoken instruction together with the target script and are asked to generate the script in speech.
WER remains nearly unchanged from the script-only condition, suggesting that spoken-instruction conditioning does not substantially impair content reproduction.

\paragraph{Visual Acoustic Cue Inference}
Finally, we examine whether models can infer target acoustic values from visual context.
To isolate this capability from speech generation, we reformulate the problem as a multiple-choice task.
Given an image associated with a feature, the model must select the target value implied by the image (e.g., \textit{Select the appropriate speech rate associated with the image provided: (1) fast (2) slow}).
Models achieve high accuracy, indicating that they can recover acoustic cues from visual scenes reliably in isolation.

Overall, the outcomes imply that the difficulty of \benchmark\ cannot be explained solely by failures in script reading, spoken-instruction conditioning, or visual cue inference. 
Instead, the main challenge potentially lies in integrating these components into context-grounded speech generation.


\subsection{Controlled Input Decomposition}
\label{subsec:task-decomp}
The original task in \benchmark\ requires a model to preserve the target script, identify the relevant acoustic dimension from the spoken instruction, infer its intended value from visual context, and realize it in generated speech.
To pinpoint where this process fails, we perform controlled input decomposition, progressively replacing non-text inputs with textual surrogates.


\begin{figure}[t]  
\centering
\includegraphics[width=1\columnwidth, keepaspectratio]{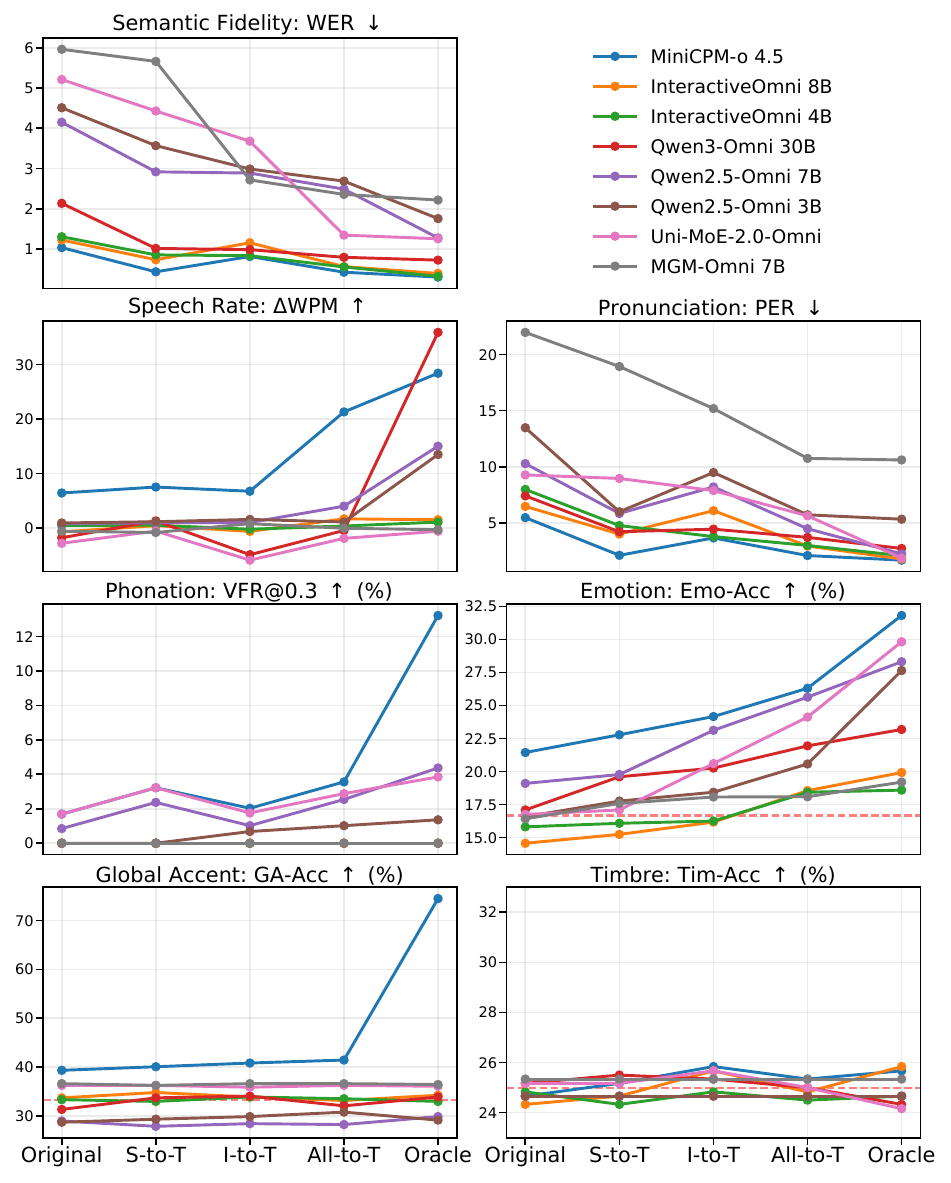}
  \caption{
Results of Controlled Input Decomposition across all evaluation metrics.
Starting from the Original setting, inputs are progressively textualized through S-to-T, I-to-T, and All-to-T, while Oracle explicitly specifies the target acoustic value.
Dashed red lines in Emo-Acc, GA-Acc, and Tim-Acc indicate random baselines.
}
\label{fig:analysis_ablation_study}
\end{figure}

Starting from the \textbf{Original} setting (spoken instruction, image, and text script), we introduce four simplified variants: \textbf{S-to-T}, \textbf{I-to-T}, \textbf{All-to-T}, and \textbf{Oracle}.
In \textbf{S-to-T}, the spoken instruction is replaced with its textual counterpart. 
In \textbf{I-to-T}, the image is replaced with descriptive keywords. 
In \textbf{All-to-T}, all inputs are textualized, removing multimodal integration while still requiring the model to infer the target acoustic value from contextual description. 
Finally, in \textbf{Oracle}, the target acoustic value is explicitly specified in the instruction (e.g., \textit{``Speak the provided Target Script Text as fast as possible.''}).\footnote{Figure~\ref{fig:all-to-T-vs-Oracle} illustrates the gap between  All-to-T and Oracle.}
This decomposition is enabled by the synthetic construction of \benchmark, as the control variables used during data generation can be directly reused as experimental conditions.

Figure~\ref{fig:analysis_ablation_study} shows that performance generally improves as the task becomes more explicit and approaches the Oracle condition.
However, both the magnitude and the shape of this improvement vary substantially across features and models, suggesting that the observed weakness cannot be reduced to a single bottleneck.



\paragraph{Failure Type I: Lack of Direct Acoustic Control}
For some features, performance remains at chance level even under Oracle conditions, indicating that the model cannot reliably realize the requested acoustic value even when no contextual inference is required.
Timbre is the clearest example: all models remain near random across all conditions.
Global Accent and Phonation show similar behavior for most models, with MiniCPM-o 4.5 as the main exception.
These results suggest that certain acoustic properties remain nearly uncontrollable in current omni-modal models.

\paragraph{Failure Type II: Failure of Implicit Acoustic Inference}
A second failure mode occurs when models can execute acoustic control once the target value is explicitly provided, but fail when it must be inferred from context.
In such cases, performance increases sharply in Oracle while remaining at chance through All-to-T.
For example, Speech Rate of Qwen3-Omni 30B, Qwen2.5-Omni 7B, and Qwen2.5-Omni 3B improves substantially when given explicit instructions, yet remains close to random when the target acoustic value must be inferred from context. 
A similar pattern appears for Phonation and Global Accent in MiniCPM-o 4.5. 
This suggests that direct acoustic control and implicit acoustic inference are separable capabilities, and that many models exhibit the former without reliably achieving the latter.

\paragraph{Failure Type III: Failure in Multimodal Acoustic Grounding}
The most challenging failure mode arises when models can infer the target acoustic value from textualized context but fail once the same information is distributed across modalities. 
MiniCPM-o 4.5 exhibits this pattern for Speech Rate. 
Its performance remains relatively strong in the All-to-T condition, indicating successful text-based inference, but drops in S-to-T and I-to-T. 
This suggests that multimodal acoustic grounding---linking contextually inferred acoustic intent to the actual acoustic realization of speech---remains a distinct and unresolved challenge, even when text-based inference and explicit control are available.

\paragraph{Emotion as a Relative Exception}
Emotion differs from the other features. 
Most models remain above chance across all conditions and improve steadily from Original to Oracle, suggesting that emotional cues are relatively easier to recover from context. 
At the same time, this contrast indicates that current omni-modal models capture emotional cues more reliably than broader acoustic properties.

\begin{table}[t]
\centering
\footnotesize
\setlength{\tabcolsep}{2.25pt}
\begin{tabular}{lcccccc}
\toprule
\multirow{2}{*}{}
& \multicolumn{2}{c}{\textbf{Em. (\%)}}
& \multicolumn{2}{c}{\textbf{GA. (\%)}}
& \multicolumn{2}{c}{\textbf{Ti. (\%)}} \\
\cmidrule(lr){2-3}
\cmidrule(lr){4-5}
\cmidrule(lr){6-7}
\textbf{Models} & E & H & E & H & E & H \\
\midrule
MiniCPM-o 4.5 (Ori.)  & 21.7 & 23.3 & 41.7 & 40.0 & 26.3 & 24.8 \\
MiniCPM-o 4.5 (Ora.)  & 34.2 & 36.5 & 73.3 & 74.0 & 25.0 & 24.5 \\
Qwen3-Omni 30B (Ori.) & 19.2 & 17.2 & 33.3 & 31.3 & 25.0 & 25.3 \\
Qwen3-Omni 30B (Ora.) & 24.2 & 22.2 & 35.0 & 34.7 & 25.0 & 25.3 \\
\bottomrule
\end{tabular}
\caption{
Human validation on a class-balanced subset of abstract features.
\textbf{E} denotes evaluator accuracy, and \textbf{H} denotes human accuracy aggregated across five annotators.
\textbf{Abbreviations:} \textbf{Em}otion, \textbf{G}lobal \textbf{A}ccent, \textbf{Ti}mbre, \textbf{Ori}ginal, \textbf{Ora}cle.
}
\label{tab:human-eval-repr-ori-ora}
\end{table}

\paragraph{Evaluator-Human Agreement}
Finally, we investigate evaluator–human agreement on model-generated speech using a class-balanced subset of Emotion, Global Accent, and Timbre under two conditions: Original and Oracle.
The subset comprises 260 benchmark instances (120/60/80 for Emotion/Global Accent/Timbre), yielding 1,040 generated clips in total across two models and two conditions, with five human annotations per clip.
As shown in Table \ref{tab:human-eval-repr-ori-ora}, evaluator accuracies closely align with human accuracies across features, models, and conditions, supporting their use as scalable proxies in subsequent experiments.

\begin{figure}[t]  
\centering
\includegraphics[width=1.0\columnwidth, keepaspectratio]{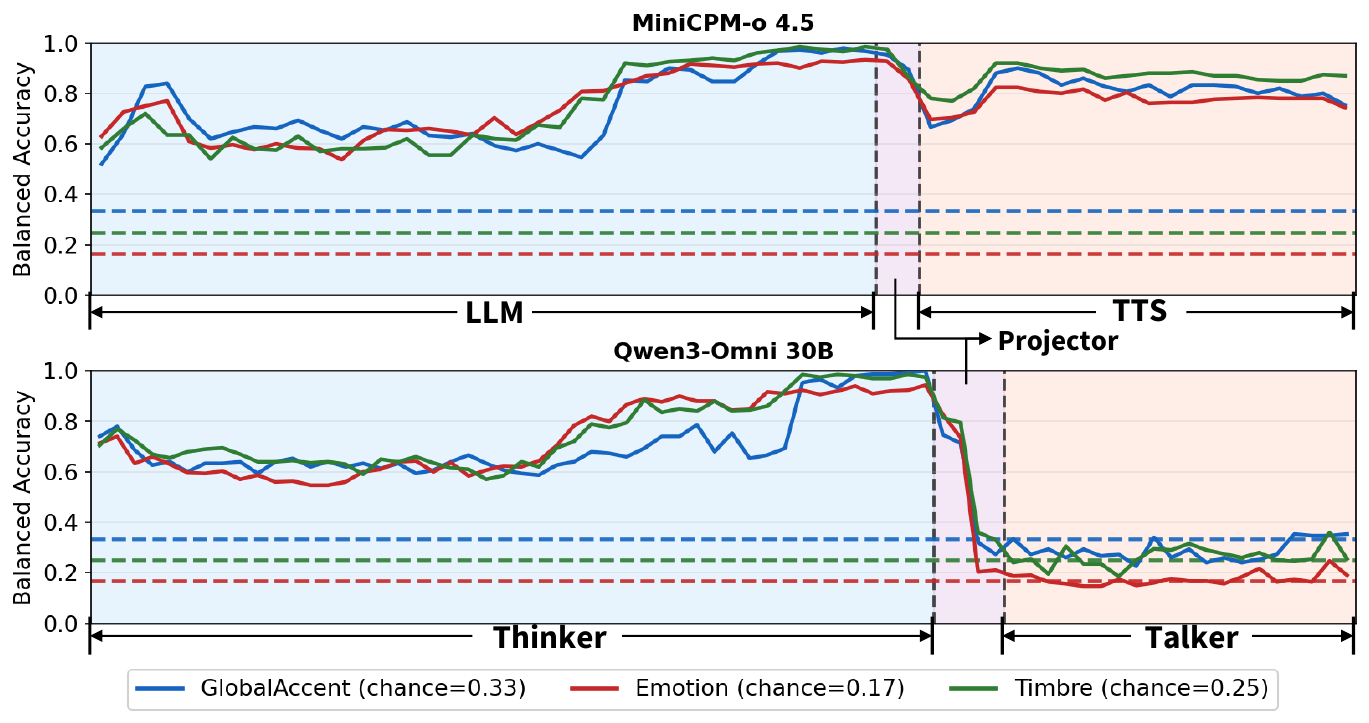}
\caption{
Linear probing of context-relevant information across model layers.
MiniCPM-o 4.5 preserves decodable context into the TTS decoder, whereas Qwen3-Omni 30B drops to near chance in the Talker.
}
\label{fig:linear-probe-mini-qwen}
\end{figure}

\subsection{Context Flow to Speech Generation}
\label{subsec:context-flow}
To better understand the performance gap on \benchmark, we compare MiniCPM-o 4.5, the best-performing model on our benchmark, with Qwen3-Omni 30B, which performs strongly on prior benchmarks but less well in this setting.
Using layer-wise hidden states, we train linear probes to predict the intended acoustic value, and use balanced accuracy to examine whether multimodal context remains linearly decodable as it propagates through each model.
Figure \ref{fig:linear-probe-mini-qwen} shows that MiniCPM-o 4.5 retains context-relevant acoustic information throughout the LLM backbone and into the TTS decoder, whereas Qwen3-Omni 30B drops from high decodability in the Thinker to chance in the Talker.
This comparison points to a possible architectural advantage of models with tighter hidden-state integration across modality-specific components and the LLM, as in MiniCPM-o 4.5, over more decoupled Thinker--Talker designs such as Qwen3-Omni 30B for context-grounded acoustic control.

\section{Conclusion}

We proposed \benchmark\ to evaluate a capability that has been largely overlooked in prior omni-modal evaluation: generating speech responses whose acoustic delivery correctly reflects multimodal context.
Experiments on the benchmark reveal that strong performance on existing omni-modal benchmarks based on textual outputs does not readily transfer to this speech-oriented setting.
The performance gap is not simply due to insufficient ability to perform individual elementary operations.
Instead, our analyses point to three sources of weakness: (1) limited direct control over certain acoustic attributes, (2) challenges in inferring implicit targets from context, and (3) unstable grounding under distributed multimodal information.
Linear probing further suggests that stronger acoustic control is associated with better retention of context-relevant information up to the speech generation stage.
We hope \benchmark\ will serve as a foundation for advancing context-grounded speech generation in omni-modal models.

\section*{Limitations}
First, speech generation may involve joint control of multiple acoustic features (e.g., a fast speech rate and angry emotion), whereas each instance in \benchmark\ evaluates only a single target acoustic feature.
As an initial testbed, we leave this compositional setting to future work, especially given that current omni-modal models still struggle even in the single-feature setting.

Second, \benchmark\ focuses on spoken instructions as the audio input, excluding non-speech signals such as environmental sounds or background music that may also provide contextual cues for acoustic delivery. Incorporating such audio contexts is left for future work.

\section*{Ethics Statement}
All instances in OmniACBench, including scripts, spoken instructions, and images, are synthetically generated, and no external copyrighted data are directly redistributed. OmniACBench is released by NAVER Cloud Corp. under the Apache License 2.0, allowing its use and redistribution in accordance with the license terms.

\section*{Acknowledgments}
This work was supported by Institute of Information \& communications Technology Planning \& Evaluation (IITP) grant funded by the Korea government(MSIT) (No.RS-2020-II201373, Artificial Intelligence Graduate School Program(Hanyang University)).
This work was supported by Institute of Information \& communications Technology Planning \& Evaluation (IITP) under the artificial intelligence semiconductor support program to nurture the best talents (IITP-2026-RS-2023-00253914) grant funded by the Korea government(MSIT).

\bibliography{custom}

@misc{gemini3proimage_modelcard,
  title        = {Gemini 3 Pro Image Generation Model Card},
  author       = {{Google}},
  year         = {2025},
  howpublished = {\url{https://storage.googleapis.com/deepmind-media/Model-Cards/Gemini-3-Pro-Image-Model-Card.pdf}},
  note         = {Accessed: 2026-03-11}
}

@misc{gemini25flashimage_modelcard,
  title        = {Gemini 2.5 Flash Image Model Card},
  author       = {{Google}},
  year         = {2025},
  howpublished = {\url{https://storage.googleapis.com/deepmind-media/Model-Cards/Gemini-2-5-Flash-Model-Card.pdf}},
  note         = {Accessed: 2026-03-11}
}

@misc{gpt5_systemcard,
  title        = {GPT-5 System Card},
  author       = {{OpenAI}},
  year         = {2025},
  howpublished = {\url{https://openai.com/index/gpt-5-system-card/}},
  note         = {Accessed: 2026-03-11}
}

@misc{gptimage1_modelcard,
  title        = {GPT-Image-1},
  author       = {{OpenAI}},
  year         = {2025},
  howpublished = {\url{https://platform.openai.com/docs/models/gpt-image-1}},
  note         = {Accessed: 2026-03-11}
}

@misc{anthropic_claude_models,
  title        = {Claude Models Overview},
  author       = {{Anthropic}},
  year         = {2025},
  howpublished = {\url{https://docs.anthropic.com/en/docs/about-claude/models/overview}},
  note         = {Accessed: 2026-03-11}
}

@misc{xai2025grok4,
  title        = {Grok 4},
  author       = {{xAI}},
  year         = {2025},
  howpublished = {\url{https://x.ai/news/grok-4}},
  note         = {Accessed: 2026-03-11}
}

@misc{openbmb2026minicpmo,
  title        = {MiniCPM-o Series},
  author       = {{OpenBMB}},
  year         = {2026},
  howpublished = {\url{https://github.com/OpenBMB/MiniCPM-o}},
  note         = {Accessed: 2026-03-11}
}

@misc{li2025unimoe20,
      title={Uni-MoE-2.0-Omni: Scaling Language-Centric Omnimodal Large Model with Advanced MoE, Training and Data}, 
      author={Yunxin Li and Xinyu Chen and Shenyuan Jiang and Haoyuan Shi and Zhenyu Liu and Xuanyu Zhang and Nanhao Deng and Zhenran Xu and Yicheng Ma and Meishan Zhang and Baotian Hu and Min Zhang},
      year={2025},
      eprint={2511.12609},
      archivePrefix={arXiv},
      primaryClass={cs.CL},
      url={https://arxiv.org/abs/2511.12609}, 
}

@misc{eleven_multilingual_v2_modeldoc,
  title        = {Eleven Multilingual v2 (Model Documentation)},
  author       = {{ElevenLabs}},
  year         = {2026},
  howpublished = {\url{https://elevenlabs.io/docs/overview/models#multilingual-v2}},
  note         = {Accessed: 2026-03-11}
}

@misc{eleven_flash_v25_modeldoc,
  title        = {Eleven Flash v2.5 (Model Documentation)},
  author       = {{ElevenLabs}},
  year         = {2026},
  howpublished = {\url{https://elevenlabs.io/docs/overview/models#flash-v25}},
  note         = {Accessed: 2026-03-11}
}

@misc{eleven_turbo_v25_modeldoc,
  title        = {Eleven Turbo v2.5 (Model Documentation)},
  author       = {{ElevenLabs}},
  year         = {2026},
  howpublished = {\url{https://elevenlabs.io/docs/overview/models#turbo-v25}},
  note         = {Accessed: 2026-03-11}
}

@article{alayrac2022flamingo,
  title={Flamingo: a visual language model for few-shot learning},
  author={Alayrac, Jean-Baptiste and Donahue, Jeff and Luc, Pauline and Miech, Antoine and Barr, Iain and Hasson, Yana and Lenc, Karel and Mensch, Arthur and Millican, Katherine and Reynolds, Malcolm and others},
  journal={Advances in neural information processing systems},
  volume={35},
  pages={23716--23736},
  year={2022}
}

@misc{li2023blip,
      title={BLIP-2: Bootstrapping Language-Image Pre-training with Frozen Image Encoders and Large Language Models}, 
      author={Junnan Li and Dongxu Li and Silvio Savarese and Steven Hoi},
      year={2023},
      eprint={2301.12597},
      archivePrefix={arXiv},
      primaryClass={cs.CV},
      url={https://arxiv.org/abs/2301.12597}, 
}

@article{deshmukh2023pengi,
  title={Pengi: An audio language model for audio tasks},
  author={Deshmukh, Soham and Elizalde, Benjamin and Singh, Rita and Wang, Huaming},
  journal={Advances in Neural Information Processing Systems},
  volume={36},
  pages={18090--18108},
  year={2023}
}

@article{tang2023salmonn,
  title={Salmonn: Towards generic hearing abilities for large language models},
  author={Tang, Changli and Yu, Wenyi and Sun, Guangzhi and Chen, Xianzhao and Tan, Tian and Li, Wei and Lu, Lu and Ma, Zejun and Zhang, Chao},
  journal={arXiv preprint arXiv:2310.13289},
  year={2023}
}

@inproceedings{han2024onellm,
  title={Onellm: One framework to align all modalities with language},
  author={Han, Jiaming and Gong, Kaixiong and Zhang, Yiyuan and Wang, Jiaqi and Zhang, Kaipeng and Lin, Dahua and Qiao, Yu and Gao, Peng and Yue, Xiangyu},
  booktitle={Proceedings of the IEEE/CVF Conference on Computer Vision and Pattern Recognition},
  pages={26584--26595},
  year={2024}
}

@article{li2024baichuan,
  title={Baichuan-omni technical report},
  author={Li, Yadong and Sun, Haoze and Lin, Mingan and Li, Tianpeng and Dong, Guosheng and Zhang, Tao and Ding, Bowen and Song, Wei and Cheng, Zhenglin and Huo, Yuqi and others},
  journal={arXiv preprint arXiv:2410.08565},
  year={2024}
}

@misc{xu2025qwen25omnitechnicalreport,
      title={Qwen2.5-Omni Technical Report}, 
      author={Jin Xu and Zhifang Guo and Jinzheng He and Hangrui Hu and Ting He and Shuai Bai and Keqin Chen and Jialin Wang and Yang Fan and Kai Dang and Bin Zhang and Xiong Wang and Yunfei Chu and Junyang Lin},
      year={2025},
      eprint={2503.20215},
      archivePrefix={arXiv},
      primaryClass={cs.CL},
      url={https://arxiv.org/abs/2503.20215}, 
}

@article{xu2025qwen3,
  title={Qwen3-omni technical report},
  author={Xu, Jin and Guo, Zhifang and Hu, Hangrui and Chu, Yunfei and Wang, Xiong and He, Jinzheng and Wang, Yuxuan and Shi, Xian and He, Ting and Zhu, Xinfa and others},
  journal={arXiv preprint arXiv:2509.17765},
  year={2025}
}

@article{wang2025mgm,
  title={MGM-Omni: Scaling Omni LLMs to Personalized Long-Horizon Speech},
  author={Wang, Chengyao and Zhong, Zhisheng and Peng, Bohao and Yang, Senqiao and Liu, Yuqi and Gui, Haokun and Xia, Bin and Li, Jingyao and Yu, Bei and Jia, Jiaya},
  journal={arXiv preprint arXiv:2509.25131},
  year={2025}
}

@article{tong2025interactiveomni,
  title={Interactiveomni: A unified omni-modal model for audio-visual multi-turn dialogue},
  author={Tong, Wenwen and Guo, Hewei and Ran, Dongchuan and Chen, Jiangnan and Lu, Jiefan and Wang, Kaibin and Li, Keqiang and Zhu, Xiaoxu and Li, Jiakui and Li, Kehan and others},
  journal={arXiv preprint arXiv:2510.13747},
  year={2025}
}

@article{li2024omnibench,
  title={Omnibench: Towards the future of universal omni-language models},
  author={Li, Yizhi and Ma, Yinghao and Zhang, Ge and Yuan, Ruibin and Zhu, Kang and Guo, Hangyu and Liang, Yiming and Liu, Jiaheng and Wang, Zekun and Yang, Jian and others},
  journal={arXiv preprint arXiv:2409.15272},
  year={2024}
}

@article{hong2025worldsense,
  title={Worldsense: Evaluating real-world omnimodal understanding for multimodal llms},
  author={Hong, Jack and Yan, Shilin and Cai, Jiayin and Jiang, Xiaolong and Hu, Yao and Xie, Weidi},
  journal={arXiv preprint arXiv:2502.04326},
  year={2025}
}

@article{zhou2025daily,
  title={Daily-omni: Towards audio-visual reasoning with temporal alignment across modalities},
  author={Zhou, Ziwei and Wang, Rui and Wu, Zuxuan},
  journal={arXiv preprint arXiv:2505.17862},
  year={2025}
}

@article{li2025omnivideobench,
  title={Omnivideobench: Towards audio-visual understanding evaluation for omni mllms},
  author={Li, Caorui and Chen, Yu and Ji, Yiyan and Xu, Jin and Cui, Zhenyu and Li, Shihao and Zhang, Yuanxing and Wang, Wentao and Song, Zhenghao and Zhang, Dingling and others},
  journal={arXiv preprint arXiv:2510.10689},
  year={2025}
}

@article{chen2025uno,
  title={UNO-Bench: A Unified Benchmark for Exploring the Compositional Law Between Uni-modal and Omni-modal in Omni Models},
  author={Chen, Chen and Hu, ZeYang and Chen, Fengjiao and Ma, Liya and Liu, Jiaxing and Li, Xiaoyu and Wang, Ziwen and Cao, Xuezhi and Cai, Xunliang},
  journal={arXiv preprint arXiv:2510.18915},
  year={2025}
}

@article{nguyen2025see,
  title={See, Hear, and Understand: Benchmarking Audiovisual Human Speech Understanding in Multimodal Large Language Models},
  author={Nguyen, Le Thien Phuc and Yu, Zhuoran and Hang, Samuel Low Yu and An, Subin and Lee, Jeongik and Ban, Yohan and Chung, SeungEun and Nguyen, Thanh-Huy and Maeng, JuWan and Lee, Soochahn and others},
  journal={arXiv preprint arXiv:2512.02231},
  year={2025}
}

@article{kim2025omhbench,
  title={OMHBench: Benchmarking Balanced and Grounded Omni-Modal Multi-Hop Reasoning},
  author={Kim, Seunghee and Bang, Ingyu and Jang, Seokgyu and Kim, Changhyeon and Bae, Sanghwan and Choi, Jihun and Xuan, Richeng and Kim, Taeuk},
  journal={arXiv preprint arXiv:2508.16198},
  year={2025}
}

@article{chen2026futureomni,
  title={FutureOmni: Evaluating Future Forecasting from Omni-Modal Context for Multimodal LLMs},
  author={Chen, Qian and Fu, Jinlan and Li, Changsong and Ng, See-Kiong and Qiu, Xipeng},
  journal={arXiv preprint arXiv:2601.13836},
  year={2026}
}

@article{guyer2021paralinguistic,
  title={Paralinguistic features communicated through voice can affect appraisals of confidence and evaluative judgments},
  author={Guyer, Joshua J and Bri{\~n}ol, Pablo and Vaughan-Johnston, Thomas I and Fabrigar, Leandre R and Moreno, Lorena and Petty, Richard E},
  journal={Journal of nonverbal behavior},
  volume={45},
  number={4},
  pages={479--504},
  year={2021},
  publisher={Springer}
}

@inproceedings{liu2022towards,
  title={Towards better characterization of paraphrases},
  author={Liu, Timothy and others},
  booktitle={Proceedings of the 60th Annual Meeting of the Association for Computational Linguistics (Volume 1: Long Papers)},
  pages={8592--8601},
  year={2022}
}

@misc{zhang2019paws,
      title={PAWS: Paraphrase Adversaries from Word Scrambling}, 
      author={Yuan Zhang and Jason Baldridge and Luheng He},
      year={2019},
      eprint={1904.01130},
      archivePrefix={arXiv},
      primaryClass={cs.CL},
      url={https://arxiv.org/abs/1904.01130}, 
}

@inproceedings{kumar2023torchaudio,
  title={Torchaudio-squim: Reference-less speech quality and intelligibility measures in torchaudio},
  author={Kumar, Anurag and Tan, Ke and Ni, Zhaoheng and Manocha, Pranay and Zhang, Xiaohui and Henderson, Ethan and Xu, Buye},
  booktitle={ICASSP 2023-2023 IEEE International Conference on Acoustics, Speech and Signal Processing (ICASSP)},
  pages={1--5},
  year={2023},
  organization={IEEE}
}

@article{li2025powsm,
  title={Powsm: A phonetic open whisper-style speech foundation model},
  author={Li, Chin-Jou and Chang, Kalvin and Bharadwaj, Shikhar and Yeo, Eunjung and Choi, Kwanghee and Zhu, Jian and Mortensen, David and Watanabe, Shinji},
  journal={arXiv preprint arXiv:2510.24992},
  year={2025}
}

@misc{radford2023robust,
      title={Robust Speech Recognition via Large-Scale Weak Supervision}, 
      author={Alec Radford and Jong Wook Kim and Tao Xu and Greg Brockman and Christine McLeavey and Ilya Sutskever},
      year={2022},
      eprint={2212.04356},
      archivePrefix={arXiv},
      primaryClass={eess.AS},
      url={https://arxiv.org/abs/2212.04356}, 
}

@article{zhao2016study,
  title={Study of the formant and duration in Chinese whispered vowel speech},
  author={Zhao, Yue and Lin, Wei},
  journal={Applied Acoustics},
  volume={114},
  pages={240--243},
  year={2016},
  publisher={Elsevier}
}

@inproceedings{gudepu2020whisper,
  title     = {{Whisper Augmented End-to-End/Hybrid Speech Recognition System — CycleGAN Approach}},
  author    = {Prithvi R.R. Gudepu and Gowtham P. Vadisetti and Abhishek Niranjan and Kinnera Saranu and Raghava Sarma and M. Ali Basha Shaik and Periyasamy Paramasivam},
  year      = {2020},
  booktitle = {{Interspeech 2020}},
  pages     = {2302--2306},
  doi       = {10.21437/Interspeech.2020-2639},
  issn      = {2958-1796},
}

@article{chen2022wavlm,
  title={Wavlm: Large-scale self-supervised pre-training for full stack speech processing},
  author={Chen, Sanyuan and Wang, Chengyi and Chen, Zhengyang and Wu, Yu and Liu, Shujie and Chen, Zhuo and Li, Jinyu and Kanda, Naoyuki and Yoshioka, Takuya and Xiao, Xiong and others},
  journal={IEEE Journal of Selected Topics in Signal Processing},
  volume={16},
  number={6},
  pages={1505--1518},
  year={2022},
  publisher={IEEE}
}

@inproceedings{dolan2005automatically,
    title = "Automatically Constructing a Corpus of Sentential Paraphrases",
    author = "Dolan, William B.  and
      Brockett, Chris",
    booktitle = "Proceedings of the Third International Workshop on Paraphrasing ({IWP}2005)",
    year = "2005",
    url = "https://aclanthology.org/I05-5002/"
}

@inproceedings{zhang2018unreasonable,
  title={The unreasonable effectiveness of deep features as a perceptual metric},
  author={Zhang, Richard and Isola, Phillip and Efros, Alexei A and Shechtman, Eli and Wang, Oliver},
  booktitle={Proceedings of the IEEE conference on computer vision and pattern recognition},
  pages={586--595},
  year={2018}
}

@misc{radford2021learning,
      title={Learning Transferable Visual Models From Natural Language Supervision}, 
      author={Alec Radford and Jong Wook Kim and Chris Hallacy and Aditya Ramesh and Gabriel Goh and Sandhini Agarwal and Girish Sastry and Amanda Askell and Pamela Mishkin and Jack Clark and Gretchen Krueger and Ilya Sutskever},
      year={2021},
      eprint={2103.00020},
      archivePrefix={arXiv},
      primaryClass={cs.CV},
      url={https://arxiv.org/abs/2103.00020}, 
}

@article{yan2025uro,
  title={URO-Bench: Towards Comprehensive Evaluation for End-to-End Spoken Dialogue Models},
  author={Yan, Ruiqi and Li, Xiquan and Chen, Wenxi and Niu, Zhikang and Yang, Chen and Ma, Ziyang and Yu, Kai and Chen, Xie},
  journal={arXiv preprint arXiv:2502.17810},
  year={2025}
}

@article{liu2025vocalbench,
  title={Vocalbench: Benchmarking the vocal conversational abilities for speech interaction models},
  author={Liu, Heyang and Wang, Yuhao and Cheng, Ziyang and Liu, Hongcheng and Li, Yiqi and Hou, Yixuan and Wu, Ronghua and Gu, Qunshan and Wang, Yanfeng and Wang, Yu},
  journal={arXiv preprint arXiv:2505.15727},
  year={2025}
}

@inproceedings{yang2025emovoice,
  title={Emovoice: Llm-based emotional text-to-speech model with freestyle text prompting},
  author={Yang, Guanrou and Yang, Chen and Chen, Qian and Ma, Ziyang and Chen, Wenxi and Wang, Wen and Wang, Tianrui and Yang, Yifan and Niu, Zhikang and Liu, Wenrui and others},
  booktitle={Proceedings of the 33rd ACM International Conference on Multimedia},
  pages={10748--10757},
  year={2025}
}

@article{nguyen2023expresso,
  title={Expresso: A benchmark and analysis of discrete expressive speech resynthesis},
  author={Nguyen, Tu Anh and Hsu, Wei-Ning and d'Avirro, Antony and Shi, Bowen and Gat, Itai and Fazel-Zarani, Maryam and Remez, Tal and Copet, Jade and Synnaeve, Gabriel and Hassid, Michael and others},
  journal={arXiv preprint arXiv:2308.05725},
  year={2023}
}

@article{jiang2025s2s,
  title={S2s-arena, evaluating speech2speech protocols on instruction following with paralinguistic information},
  author={Jiang, Feng and Lin, Zhiyu and Bu, Fan and Du, Yuhao and Wang, Benyou and Li, Haizhou},
  journal={arXiv preprint arXiv:2503.05085},
  year={2025}
}

@article{yang2025paras2s,
  title={ParaS2S: Benchmarking and Aligning Spoken Language Models for Paralinguistic-aware Speech-to-Speech Interaction},
  author={Yang, Shu-wen and Tu, Ming and Liu, Andy T and Qu, Xinghua and Lee, Hung-yi and Lu, Lu and Wang, Yuxuan and Wu, Yonghui},
  journal={arXiv preprint arXiv:2511.08723},
  year={2025}
}

@article{wang2025voiceassistant,
  title={VoiceAssistant-Eval: Benchmarking AI Assistants across Listening, Speaking, and Viewing},
  author={Wang, Ke and Ren, Houxing and Lu, Zimu and Zhan, Mingjie and Li, Hongsheng},
  journal={arXiv preprint arXiv:2509.22651},
  year={2025}
}

@article{cao2014crema,
  title={Crema-d: Crowd-sourced emotional multimodal actors dataset},
  author={Cao, Houwei and Cooper, David G and Keutmann, Michael K and Gur, Ruben C and Nenkova, Ani and Verma, Ragini},
  journal={IEEE transactions on affective computing},
  volume={5},
  number={4},
  pages={377--390},
  year={2014},
  publisher={IEEE}
}

@article{dupuis2011recognition, 
title={Recognition of emotional speech for younger and older talkers: Behavioural findings from the toronto emotional speech set}, 
volume={39}, 
url={https://jcaa.caa-aca.ca/index.php/jcaa/article/view/2471}, 
abstractNote={A study that was conducted to analyze recognition of emotional speech for younger and older talkers is presented. Each actor recorded the stimuli individually in a sound- attenuating booth for approximately 20 hours. During the recording sessions, which typically lasted three to four hours, the majority of the time was spent creating the voice recordings, while approximately 10% of the time was devoted to practicing and fine-tuning each actor’s portrayal of each of the emotions. Three female undergraduate students with normal hearing listened to the stimuli and identified, for each actor, which token of each NU-6 item they considered to be the most representative for each of the seven emotions. The experimenter used the same procedure to listen to each of the sound files.}, 
number={3}, 
journal={Canadian Acoustics}, 
author={Dupuis, Kate and Kathleen Pichora-Fuller, M.}, 
year={2011}, 
month={Sep.}, 
pages={182–183} }

@article{livingstone2018ryerson,
  title={The Ryerson Audio-Visual Database of Emotional Speech and Song (RAVDESS): A dynamic, multimodal set of facial and vocal expressions in North American English},
  author={Livingstone, Steven R and Russo, Frank A},
  journal={PloS one},
  volume={13},
  number={5},
  pages={e0196391},
  year={2018},
  publisher={Public Library of Science}
}

@incollection{haq2011multimodal,
  title={Multimodal emotion recognition},
  author={Haq, Sanaul and Jackson, Philip JB},
  booktitle={Machine audition: principles, algorithms and systems},
  pages={398--423},
  year={2011},
  publisher={IGI Global Scientific Publishing}
}

@article{wang2024globe,
  title={Globe: A high-quality english corpus with global accents for zero-shot speaker adaptive text-to-speech},
  author={Wang, Wenbin and Song, Yang and Jha, Sanjay},
  journal={arXiv preprint arXiv:2406.14875},
  year={2024}
}

@inproceedings{sanabria2023edinburgh,
  title={The edinburgh international accents of english corpus: Towards the democratization of english asr},
  author={Sanabria, Ramon and Bogoychev, Nikolay and Markl, Nina and Carmantini, Andrea and Klejch, Ondrej and Bell, Peter},
  booktitle={ICASSP 2023-2023 IEEE International Conference on Acoustics, Speech and Signal Processing (ICASSP)},
  pages={1--5},
  year={2023},
  organization={IEEE}
}

@inproceedings{gerz2021multilingual,
  title={Multilingual and cross-lingual intent detection from spoken data},
  author={Gerz, Daniela and Su, Pei-Hao and Kusztos, Razvan and Mondal, Avishek and Lis, Micha{\l} and Singhal, Eshan and Mrk{\v{s}}i{\'c}, Nikola and Wen, Tsung-Hsien and Vuli{\'c}, Ivan},
  booktitle={Proceedings of the 2021 Conference on Empirical Methods in Natural Language Processing},
  pages={7468--7475},
  year={2021}
}

@misc{commonaccent,
  author       = {{DTU54DL}},
  title        = {common-accent},
  howpublished = {\url{https://huggingface.co/datasets/DTU54DL/common-accent}},
  note         = {Hugging Face dataset card, accessed March 13, 2026}
}

@misc{gemini3flash_modelcard,
  title        = {Gemini 3 Flash Model Card},
  author       = {{Google}},
  year         = {2025},
  howpublished = {\url{https://storage.googleapis.com/deepmind-media/Model-Cards/Gemini-3-Flash-Model-Card.pdf}},
  note         = {Accessed: 2026-03-01}
}

@inproceedings{ma2024emotion2vec,
  title={emotion2vec: Self-supervised pre-training for speech emotion representation},
  author={Ma, Ziyang and Zheng, Zhisheng and Ye, Jiaxin and Li, Jinchao and Gao, Zhifu and Zhang, Shiliang and Chen, Xie},
  booktitle={Findings of the Association for Computational Linguistics: ACL 2024},
  pages={15747--15760},
  year={2024}
}

@misc{zuluaga_commonaccent_ecapa_hf,
  author = {Juan Pablo Zuluaga},
  title = {Accent Identification from Speech Recordings with ECAPA-TDNN embeddings on CommonAccent},
  year = {2023},
  publisher = {HuggingFace},
  url = {https://huggingface.co/Jzuluaga/accent-id-commonaccent_ecapa}
}

@misc{accent-classifier-6class,
  author = {Miles Purvis},
  title = {English Accent Classifier (6 Classes)},
  year = {2025},
  publisher = {HuggingFace},
  url = {https://huggingface.co/MilesPurvis/english-accent-classifier}
}

@inproceedings{kim2025fcmr,
  title={Fcmr: robust evaluation of financial cross-modal multi-hop reasoning},
  author={Kim, Seunghee and Kim, Changhyeon and Kim, Taeuk},
  booktitle={Proceedings of the 63rd Annual Meeting of the Association for Computational Linguistics (Volume 1: Long Papers)},
  pages={23352--23380},
  year={2025}
}

@inproceedings{chen2022hilvoice,
  title={Hilvoice: Human-in-the-loop style selection for elder-facing speech synthesis},
  author={Chen, Xueyuan and Huang, Qiaochu and Wu, Xixin and Wu, Zhiyong and Meng, Helen},
  booktitle={2022 13th International Symposium on Chinese Spoken Language Processing (ISCSLP)},
  pages={86--90},
  year={2022},
  organization={IEEE}
}

@article{wingfield1999effects,
  title={Effects of age and passage difficulty on listening-rate preferences for time-altered speech},
  author={Wingfield, Arthur and Ducharme, Julie L},
  journal={The Journals of Gerontology Series B: Psychological Sciences and Social Sciences},
  volume={54},
  number={3},
  pages={P199--P202},
  year={1999},
  publisher={The Gerontological Society of America}
}

@article{janicki2015older,
  title={Why Do Older Adults Prefer Some Radio Stations? Helping to Increase Speech Understanding.},
  author={Janicki, Artur and Szczypiorski, Krzysztof},
  journal={J. Commun.},
  volume={10},
  number={11},
  pages={926--931},
  year={2015}
}

@misc{wiktionary,
  author       = {{Wiktionary contributors}},
  title        = {Wiktionary, the free dictionary},
  year         = {2026},
  howpublished = {\url{https://en.wiktionary.org/wiki/Wiktionary:Main_Page}},
  note         = {Accessed: 2026-03-11}
}

\clearpage
\appendix

\section{Additional Acoustic Feature Candidates}
\label{app:additional_acoustic_features}

The design of our construction framework also accommodates other forms of acoustic control that can be grounded in multimodal context and evaluated acoustically.

For example, \textbf{Volume} (e.g., loud vs.\ quiet) provides one such possibility: visual contexts such as crowded stadiums or quiet indoor environments can suggest the intended volume, while its realization can be assessed using signal-energy measures such as RMS.
Another example is \textbf{Prosodic Prominence}, where the visual context can indicate a word in the script to be emphasized, and the resulting prominence can be assessed using word-level acoustic cues such as F0, energy, and duration.

These examples illustrate other acoustic dimensions that can be instantiated within the same general framework.

\section{Benchmark Construction Details}
For \textbf{Speech Rate}, we define two target values, \textit{fast} and \textit{slow}.
Image keywords for \textit{fast} are built around situations that naturally imply urgency or emergency, such as \textit{a brick falling toward a construction worker} or \textit{a boiling-over pot}.
In contrast, image keywords for \textit{slow} are designed around elderly-directed interactions, such as scenes involving an \textit{elderly person}.
This was determined based on prior studies suggesting that speech directed to older adults should be delivered at a slower rate \cite{wingfield1999effects, janicki2015older, chen2022hilvoice}.

For \textbf{Pronunciation}, we focus on English heteronyms, where the same orthographic form has different pronunciations depending on meaning and context.
The reference pronunciations were determined based on the entries provided in Wiktionary \cite{wiktionary}.

For \textbf{Phonation}, we focus on the target value \textit{whisper}.
The corresponding image keywords describe situations in which whispering is pragmatically appropriate, including \textit{people studying in a library reading room} or \textit{a quiet art museum with visitors walking slowly}.
These scenes provide natural contextual grounding for suppressed vocal intensity and reduced voicing, making whisper-like phonation visually inferable.

For \textbf{Emotion}, we use six target values: \textit{joy}, \textit{surprised}, \textit{angry}, \textit{disgust}, \textit{fear}, and \textit{sadness}.
The corresponding image keywords are centered on visually salient facial affective cues, such as \textit{a person with an angry expression} or \textit{a person with a disgusted expression}.

For \textbf{Global Accent}, we consider three target values: \textit{India}, \textit{UK}, and \textit{Australia}.
We construct image keywords not only from national symbols such as flags but also from culturally and geographically distinctive visual cues associated with each country.
For example, keywords include \textit{the national flag of India} or \textit{the Opera House in Sydney, Australia}.
This strategy increases visual diversity and encourages models to infer the relevant national context from broader cultural grounding rather than from a single canonical cue.

For \textbf{Timbre}, we define four target values: \textit{adult male}, \textit{adult female}, \textit{elderly male}, and \textit{elderly female}.
Image keywords are constructed to foreground visually recognizable cues of speaker age and gender, such as \textit{an elderly woman}.

\section{Quality Control Protocol Details}
\label{app:detail-qcp}
\subsection{Data Filtering}
To ensure that \benchmark\ instances faithfully reflect the intended benchmark design, we apply three filtering criteria at different stages of the construction pipeline: \textbf{Semantic Preservation}, \textbf{Text Neutrality}, and \textbf{Image--Keyword Alignment}.
These criteria are designed to remove artifacts that could otherwise undermine the validity of the evaluation, such as semantically drifted instructions, scripts that leak the target acoustic value, or images that no longer reflect their source concepts.
For each criterion, we first conduct LLM-based filtering as a scalable first-pass screening step, and then perform human verification using the same criterion to confirm the final decision.

\textbf{Semantic Preservation} is applied to spoken instructions generated through paraphrasing.
Its purpose is to ensure that each paraphrased instruction remains faithful to the meaning and intent of the original template, without introducing semantic drift, unintended constraints, or changes in the target acoustic dimension.
This step is important because the spoken instruction serves as the control signal for the task; if its meaning changes during paraphrasing, the resulting instance may no longer represent the intended evaluation condition.
The prompt used for this filtering step is shown in Figure \ref{app-fig:filter-semantic}.

\textbf{Text Neutrality} is applied to generated scripts in order to prevent shortcut solutions.
Specifically, we remove scripts whose lexical content alone reveals the intended target value, since such cases would allow models to recover the desired acoustic realization directly from the text rather than from multimodal context.
This filtering criterion helps preserve the intended role assignment of the benchmark, where the text provides only the verbal content to be spoken and does not itself encode the acoustic target.
The prompt used for this filtering step is shown in Figure \ref{app-fig:filter-text-neutrality}.

\textbf{Image--Keyword Alignment} is applied to generated images to verify that they remain semantically consistent with the original image keywords used during construction.
Since images are synthesized from expanded prompts rather than directly from the initial keywords, this step checks whether the final image still preserves the intended scene or concept associated with the target acoustic value.
This is necessary to ensure that the visual modality provides valid contextual grounding, rather than introducing irrelevant or misleading content.
The prompt used for this filtering step is shown in Figure \ref{app-fig:filter-image-keyword}.

\subsection{Quantitative Verification}
\label{app:quanti-verify}

\paragraph{Paraphrasing Quality}
Because spoken instructions in \benchmark\ are generated by paraphrasing feature-level templates, it is important to verify that they exhibit sufficient linguistic diversity rather than remaining close to a small set of fixed phrasings.
Following prior benchmark construction work \cite{kim2025fcmr, kim2025omhbench}, we measure this property using Word Position Deviation (WPD) and Lexical Deviation (LD) \cite{liu2022towards}.
\benchmark\ achieves WPD/LD scores of 0.11/0.69, compared to 0.12/0.42 on MRPC \cite{dolan2005automatically} and 0.07/0.13 on PAWS \cite{zhang2019paws}.
These results indicate that the spoken instructions in \benchmark\ preserve substantial lexical diversity while avoiding overly rigid template repetition.

\paragraph{Speech Quality}
We also verify the quality of synthesized speech, since unintelligible or text-inaccurate audio would introduce noise unrelated to the intended benchmark capability.
To this end, we evaluate transcription fidelity using WER and CER computed with Whisper-large-v3 \cite{radford2023robust}, and intelligibility using STOI ($\uparrow$) \cite{kumar2023torchaudio}.
The resulting scores are WER 0.004, CER 0.001, and STOI 0.994, indicating that the synthesized speech remains highly faithful to the intended script while also being near-perfect in intelligibility.

\paragraph{Diversity Control}
We first evaluate whether meta-prompting improves image diversity over na\"ive keyword prompting.
For each keyword group, we compare images generated by the two approaches using average pairwise CLIP embedding cosine distance \cite{radford2021learning} and LPIPS \cite{zhang2018unreasonable}, which capture semantic and perceptual variation, respectively.
The meta-prompt approach yields higher diversity on both metrics (CLIP 0.1242 vs. 0.0671; LPIPS 0.4661 vs. 0.3733), showing that richer prompt expansion leads to greater visual variation than keyword-only prompting.
A qualitative comparison of images generated by the two approaches for the same keyword is provided in Figure \ref{fig:keyword-metaprompt}.

Second, to reduce model-specific bias and increase generation diversity, we maintain model pools for each stage of tri-modal generation and randomly sample one model per instance from the corresponding pool.
For text generation, paraphrasing, and image prompt generation, we employ three LLMs: Gemini 3 Flash \citep{gemini3flash_modelcard}, GPT 5 Nano \citep{gpt5_systemcard}, and Claude Haiku 4.5 \citep{anthropic_claude_models}.
For image generation, we utilize four models: Gemini 3 Pro Image \citep{gemini3proimage_modelcard}, Gemini 2.5 Flash Image \citep{gemini25flashimage_modelcard}, GPT Image 1, and GPT Image 1.5 \citep{gptimage1_modelcard}.
For TTS, we use three models: Eleven Multilingual v2 \citep{eleven_multilingual_v2_modeldoc}, Eleven Flash v2.5 \citep{eleven_flash_v25_modeldoc}, and Eleven Turbo v2.5 \citep{eleven_turbo_v25_modeldoc}.
At each generation stage, one model is randomly selected from the corresponding pool to increase diversity in the generated text, images, and speech.
An independent model, Grok 4 Fast \citep{xai2025grok4}, is used for LLM-based filtering.

\section{Evaluator Training Details}
\label{app:appendix-evaluator-train-detail}
Rather than relying on off-the-shelf evaluators, we train task-specific classifiers for all three abstract features.
This choice is motivated by both label-space mismatch and empirical performance gaps.
On our held-out test sets for Emotion, Global Accent, and Timbre, off-the-shelf alternatives do not consistently align with the target categories or provide sufficient accuracy: emotion2vec \cite{ma2024emotion2vec} achieves 84.57\% on Emotion, while ECAPA-TDNN-based English accent classifier \cite{zuluaga_commonaccent_ecapa_hf} and Wav2Vec2 based English accent classifier \cite{accent-classifier-6class} achieve 51.83\% and 59.91\%, respectively, on Global Accent, all below our task-specific evaluators.
For Timbre, moreover, we are not aware of a readily available off-the-shelf evaluator that directly matches our four-way setting (\textit{adult/elderly} $\times$ \textit{female/male}).
Task-specific training therefore provides a better label match and a more reproducible evaluation pipeline for \benchmark.

\paragraph{Data Collection}
We train task-specific classifiers for three abstract acoustic features: Emotion, Global Accent, and Timbre.
For \textbf{Emotion}, we aggregate speech data from four publicly available datasets---CREMA-D~\cite{cao2014crema} (6,355 samples), TESS~\cite{dupuis2011recognition} (2,400), RAVDESS~\cite{livingstone2018ryerson} (1,888), and SAVEE~\cite{haq2011multimodal} (360)---yielding 11,003 samples in total across six classes (\textit{angry, disgust, fear, joy, sadness, surprised}).
For \textbf{Global Accent}, we collect 12,000 samples from GLOBE~\cite{wang2024globe} (7,334), EDACC~\cite{sanabria2023edinburgh} (2,004), Common-Accent~\cite{commonaccent} (1,497), and MINDS-14~\cite{gerz2021multilingual} (1,165), covering three classes (\textit{Australian, Indian, UK}).
For \textbf{Timbre}, we collect 12,000 samples from GLOBE~\cite{wang2024globe} (11,400) and EDACC~\cite{sanabria2023edinburgh} (600), covering four classes (\textit{adult female, adult male, elderly female, elderly male}), where \textit{adult} refers to speakers in their 20s--30s and \textit{elderly} refers to speakers aged 60 and above.
All three datasets are split into train/dev/test partitions.

\paragraph{Training}
All classifiers use WavLM-Large~\cite{chen2022wavlm} as the backbone, with a task-specific classification head on top of mean-pooled frame representations.
We fully fine-tune the backbone using AdamW ($\beta_1=0.9$, $\beta_2=0.999$, weight decay $10^{-4}$), a cosine learning-rate schedule with 2-epoch linear warmup, label smoothing of 0.1, and gradient clipping at norm 1.0.
Training runs for 30 epochs with a per-GPU batch size of 32.
We select the final configuration via grid search over head learning rate $\in \{10^{-5}, 10^{-4}, 10^{-3}\}$ and backbone learning rate $\in \{5 \times 10^{-7}, 5 \times 10^{-6}, 5 \times 10^{-5}\}$, choosing the checkpoint with the best development-set accuracy.
The selected learning rates are $10^{-4}$ for the head and $5 \times 10^{-5}$ for the backbone for all three classifiers.
The resulting checkpoints achieve accuracies of 89.43\%, 97.29\%, and 96.67\% on the held-out test sets for Emotion, Global Accent, and Timbre, respectively, supporting their use as automatic evaluators in \benchmark.

\section{Context Flow Analysis Details}
For the context-flow analysis in Section \ref{subsec:context-flow}, we compared MiniCPM-o 4.5 \cite{openbmb2026minicpmo} and Qwen3-Omni 30B \cite{xu2025qwen3} to examine whether information about the intended acoustic attribute remains decodable as it propagates through each model toward speech generation.
In MiniCPM-o 4.5, representations were extracted from the LLM backbone, the intermediate projection stage, and the TTS decoder; in Qwen3-Omni 30B, they were extracted from the Thinker, the intermediate projection stages, and the Talker.
We then trained a separate linear probe for each layer and each acoustic feature---Emotion, Global Accent, and Timbre---to predict the target sub-category from the corresponding hidden representation.
Each probe was implemented as a single linear classifier trained with cross-entropy loss and AdamW.
Evaluation was conducted using repeated stratified train/test splits, with hidden representations standardized using training-set statistics only.
Performance was measured using balanced accuracy (i.e., mean per-class recall) on the held-out split, averaged across repeated runs.

\section{Model Details}
\label{app:model_details}

\subsection{Open-source Omni Models}
We evaluate eight omni models that support text, image, and speech inputs and generate speech outputs.
This section focuses on architecture-level characteristics relevant to context-grounded acoustic control.

\paragraph{MiniCPM-o 4.5 \citep{openbmb2026minicpmo}.}
The MiniCPM-o family adopts an end-to-end omni design that directly connects modality encoders and decoders with a language backbone through hidden-state interactions, rather than a cascaded ASR$\rightarrow$LLM$\rightarrow$TTS stack.
Its architecture combines SigLip2 for vision, Whisper-medium for audio understanding, CosyVoice2-style speech tokenization and Token2Wav, and a Qwen3-8B backbone, together with full-duplex streaming and timeline modeling (TDM) for real-time interaction.
For speech specifically, input audio is encoded online and output speech is generated by an interleaved text--speech token decoder, enabling synchronized full-duplex response generation.

\paragraph{InteractiveOmni \citep{tong2025interactiveomni}.}
InteractiveOmni integrates a vision encoder, an audio encoder, a language model, and a speech decoder into a unified architecture for audio-visual multi-turn dialogue.
A central design element is multi-stage training with dedicated data curation for long-horizon conversational context and speech-oriented response quality.
Its speech pathway is explicitly end-to-end: audio input is handled by the unified audio encoder, while speech output is produced by the integrated speech decoder rather than by an external TTS post-processor.

\paragraph{Qwen3-Omni 30B \citep{xu2025qwen3}.}
Qwen3-Omni 30B uses a Thinker--Talker Mixture-of-Experts architecture that separates high-level reasoning from speech generation.
For speech input, it adopts AuT, a lightweight continuous audio encoder with a 12.5Hz frame rate that maps audio into semantic features for the Thinker.
For speech output, the Talker autoregressively predicts multi-codebook discrete speech codec tokens and reconstructs waveform with a lightweight causal ConvNet (Code2Wav), targeting low first-packet latency and streaming generation.

\paragraph{Qwen2.5-Omni \citep{xu2025qwen25omnitechnicalreport}.}
Qwen2.5-Omni introduces an end-to-end omni stack with block-wise audio and video streaming encoders and temporal multimodal positional encoding (TM-RoPE) for synchronized time modeling.
It also adopts a Thinker--Talker paradigm: speech input is consumed through the block-wise streaming audio encoder, while speech output is produced as discrete speech tokens via a dual-track autoregressive Talker and then rendered through a sliding-window DiT-based codec pathway for streaming audio.

\paragraph{Uni-MoE-2.0-Omni \citep{li2025unimoe20}.}
Uni-MoE-2.0-Omni emphasizes scalable multimodal routing via a dynamic-capacity MoE design with shared, routed, and null experts, coupled with omni-modality 3D-RoPE for unified spatiotemporal representation.
Its training pipeline includes progressive stages from omni pretraining to preference optimization.
For speech handling, the architecture includes a unified speech encoder for audio feature extraction and dedicated speech-generation tokens that synchronize speech and text generation, combined with a context-aware MoE-TTS stage for high-quality synthesis.

\paragraph{MGM-Omni 7B \citep{wang2025mgm}.}
MGM-Omni-7B proposes a brain--mouth dual-track token architecture that decouples reasoning from low-latency speech generation.
In this design, speech input is processed by a dual audio encoder for robust understanding, while speech output is generated by chunk-based parallel decoding to narrow the text/speech token-rate gap and improve real-time responsiveness.

\subsection{Commercial Models}
For diversity control in Section \ref{subsubsec:quantitative-verification}, we use commercial model pools at each generation stage.
Since full implementation details are often undisclosed for closed models, we summarize publicly documented architectural signals and distinctive characteristics.

\paragraph{LLMs for text generation, paraphrasing, and meta-prompting.}
Gemini 3 Flash \citep{gemini3flash_modelcard} is a Gemini 3 family model with configurable thinking budgets and long-context multimodal input.
GPT 5 Nano \citep{gpt5_systemcard} belongs to the GPT 5 family with explicit fast/thinking model routing and a small reasoning variant (\texttt{gpt-5-thinking-nano}).
Claude Haiku 4.5 \citep{anthropic_claude_models} is positioned as a low-latency model in the Claude 4.5 line with extended-thinking mode and large-context handling.

\paragraph{Image generation models.}
Gemini 3 Pro Image \citep{gemini3proimage_modelcard} and Gemini 2.5 Flash Image \citep{gemini25flashimage_modelcard} provide native image generation and editing within Gemini multimodal models; Gemini 2.5 Flash additionally adopts hybrid reasoning and sparse-MoE-based scaling.
GPT Image 1 and GPT Image 1.5 \citep{gptimage1_modelcard} are OpenAI's natively multimodal image-generation models supporting text-and-image conditioning and iterative editing workflows.

\paragraph{Text-to-speech models.}
Eleven Multilingual v2 \citep{eleven_multilingual_v2_modeldoc} focuses on expressive multilingual speech quality.
Eleven Flash v2.5 \citep{eleven_flash_v25_modeldoc} targets very low latency for realtime use, while Eleven Turbo v2.5 \citep{eleven_turbo_v25_modeldoc} balances speed and quality for general-purpose synthesis.

\paragraph{LLM-based filtering model.}
We use Grok 4 Fast \citep{xai2025grok4} as a separate filtering model.
The Grok 4 family is characterized by large-scale reinforcement learning and native tool-use integration, which makes it suitable for independent quality filtering without sharing generation-stage model biases.

\section{Experimental Environment}
\label{app:appendix-exp-environ}

All experiments were conducted on a machine equipped with Intel Xeon Gold 6338 CPU @ 2.00GHz (2 sockets $\times$ 32 cores, up to 3.20GHz boost), and 4 $\times$ NVIDIA A100-SXM4 GPUs each with 80 GB of memory. The system ran Ubuntu 20.04.5 LTS with CUDA compilation tools release 11.8.
During both dataset generation and evaluation, the random seed was fixed to 42 to ensure reproducibility.


\begin{figure}[h!]
    \centering
    \begin{minipage}{0.95\columnwidth}
    \fbox{
        \begin{minipage}{0.93\columnwidth}
            \fontsize{10}{10}\selectfont
            \textbf{Text Generation Prompt} \\
            \rule{\textwidth}{0.4pt}
            \fontsize{9}{10}\selectfont  
          Your task is to generate a single line of dialogue.\\
          \\
          1. The line must be emotionally neutral.\\
          2. It must be nationally neutral.\\
          3. It must be neutral with respect to gender and age.\\
          4. The line must be a declarative sentence (not a question or exclamation).\\
          \\
          Output only the single line of dialogue with no additional text.
            \vspace{4pt}
        \end{minipage}
        }
    \end{minipage}
    
    \caption{A prompt used for text transcript generation.}
    \label{app-fig:text-transcript-prompt}
\end{figure}

\begin{figure}[h!]
    \centering
    \begin{minipage}{0.95\columnwidth}
    \fbox{
        \begin{minipage}{0.93\columnwidth}
            \fontsize{10}{10}\selectfont
            \textbf{Instruction Paraphrasing Prompt} \\
            \rule{\textwidth}{0.4pt}
            \fontsize{9}{10}\selectfont  
          Paraphrase the provided instruction template.\\
          \\
          Condition 1: The original and the paraphrased instruction must have exactly the same essential meaning.\\
          Condition 2: Perform paraphrasing with consideration of lexical variation.\\
          Condition 3: Perform paraphrasing with consideration of syntactic variation.\\
          \\
          Output only the single paraphrased instruction with no additional text.\\
          \\
          original instruction template: \{original\_template\}\\
          \\
          paraphrased instruction:
            \vspace{4pt}
        \end{minipage}
        }
    \end{minipage}
    
    \caption{A prompt used for instruction paraphrasing.}
    \label{app-fig:paraphrase-prompt}
\end{figure}

\begin{figure}[h!]
    \centering
    \begin{minipage}{0.95\columnwidth}
    \fbox{
        \begin{minipage}{0.93\columnwidth}
            \fontsize{10}{10}\selectfont
            \textbf{Image Meta-Prompt} \\
            \rule{\textwidth}{0.4pt}
            \fontsize{9}{10}\selectfont  
          You are a professional prompt engineer who writes prompts for image generation models.\\
          \\
          Create a single image generation prompt that satisfies all of the following conditions:\\
          \\
          1. The element \textbf{\{image\_keyword\}} must be the central focus of the image and be strongly emphasized and clearly visible in the prompt.\\
          2. The final output must be only one image generation prompt and nothing else.\\
          3. All elements must be separated by commas.\\
          4. The prompt must consist of 5 to 8 comma-separated elements that are visually clear and specific.\\
          \\
          Generate the image generation prompt that meets these conditions.
            \vspace{4pt}
        \end{minipage}
        }
    \end{minipage}
    
    \caption{A meta-prompt template used for image generation prompt expansion.}
    \label{app-fig:meta-prompt}
\end{figure}

\begin{figure}[h!]
    \centering
    \begin{minipage}{0.95\columnwidth}
    \fbox{
        \begin{minipage}{0.93\columnwidth}
            \fontsize{10}{10}\selectfont
            \textbf{Semantic Preservation Filtering Prompt} \\
            \rule{\textwidth}{0.4pt}
            \fontsize{9}{10}\selectfont  
          original instruction: \{original\}\\
          paraphrased instruction: \{paraphrased\}\\
          If the original instruction and the paraphrased instruction have the same meaning, output Yes; otherwise, output No.\\
          Output only Yes or No without any additional explanation.\\
          Answer:
            \vspace{4pt}
        \end{minipage}
        }
    \end{minipage}
    
    \caption{A prompt used for LLM-based semantic preservation filtering.}
    \label{app-fig:filter-semantic}
\end{figure}

\begin{figure}[h!]
    \centering
    \begin{minipage}{0.95\columnwidth}
    \fbox{
        \begin{minipage}{0.93\columnwidth}
            \fontsize{10}{10}\selectfont
            \textbf{Text Neutrality Filtering Prompt} \\
            \rule{\textwidth}{0.4pt}
            \fontsize{9}{10}\selectfont  
          transcript: \{transcript\}\\
          Target Acoustic Feature: \{Target\_Acoustic\_Feature\}\\
          Can the Target Acoustic Feature be fully inferred using only the provided transcript? Answer Yes or No.\\
          Output only Yes or No without any additional explanation.\\
          Answer:
            \vspace{4pt}
        \end{minipage}
        }
    \end{minipage}
    
    \caption{A prompt used for LLM-based text neutrality filtering.}
    \label{app-fig:filter-text-neutrality}
\end{figure}

\begin{figure}[h!]
    \centering
    \begin{minipage}{0.95\columnwidth}
    \fbox{
        \begin{minipage}{0.93\columnwidth}
            \fontsize{10}{10}\selectfont
            \textbf{Image-Keyword Alignment Filtering Prompt} \\
            \rule{\textwidth}{0.4pt}
            \fontsize{9}{10}\selectfont  
          Does the provided image depict ``\{image\_keyword\}''? Answer Yes or No.\\
          Output only Yes or No without any additional explanation.\\
          Answer:
            \vspace{4pt}
        \end{minipage}
        }
    \end{minipage}
    
    \caption{A prompt used for LLM-based image-keyword alignment filtering.}
    \label{app-fig:filter-image-keyword}
\end{figure}

\begin{figure}[h!]
    \centering
    \begin{minipage}{0.95\columnwidth}
    \fbox{
        \begin{minipage}{0.93\columnwidth}
            \fontsize{10}{10}\selectfont
            \textbf{Human Verification Instruction: Semantic Preservation} \\
            \rule{\textwidth}{0.4pt}
            \fontsize{9}{10}\selectfont  
          original instruction: \{original\}\\
          paraphrased instruction: \{paraphrased\}\\
          Do the original instruction and the paraphrased instruction have the same meaning? Answer Yes or No.
            \vspace{4pt}
        \end{minipage}
        }
    \end{minipage}
    \caption{An instruction used for human-based semantic preservation filtering.}
    \label{app-fig:human-filter-semantic}
\end{figure}

\begin{figure}[h!]
    \centering
    \begin{minipage}{0.95\columnwidth}
    \fbox{
        \begin{minipage}{0.93\columnwidth}
            \fontsize{10}{10}\selectfont
            \textbf{Human Verification Instruction: Text Neutrality} \\
            \rule{\textwidth}{0.4pt}
            \fontsize{9}{10}\selectfont  
          transcript: \{transcript\}\\
          Target Acoustic Feature: \{Target\_Acoustic\_Feature\}\\
          Can the Target Acoustic Feature be fully inferred using only the provided transcript? Answer Yes or No.
            \vspace{4pt}
        \end{minipage}
        }
    \end{minipage}
    \caption{An instruction used for human-based text neutrality filtering.}
    \label{app-fig:human-filter-text-neutrality}
\end{figure}

\begin{figure}[h!]
    \centering
    \begin{minipage}{0.95\columnwidth}
    \fbox{
        \begin{minipage}{0.93\columnwidth}
            \fontsize{10}{10}\selectfont
            \textbf{Human Verification Instruction: Image-Keyword Alignment} \\
            \rule{\textwidth}{0.4pt}
            \fontsize{9}{10}\selectfont  
          Does the provided image depict ``\{image\_keyword\}''? Answer Yes or No.
            \vspace{4pt}
        \end{minipage}
        }
    \end{minipage}
    \caption{An instruction used for human-based image-keyword alignment filtering.}
    \label{app-fig:human-filter-image-keyword}
\end{figure}

\begin{figure}[h!]
    \centering
    \begin{minipage}{0.95\columnwidth}
    \fbox{
        \begin{minipage}{0.93\columnwidth}
            \fontsize{10}{10}\selectfont
            \textbf{Human Annotator Instruction: Emotion} \\
            \rule{\textwidth}{0.4pt}
            \fontsize{9}{10}\selectfont  
            Speech: \{speech\}\\
            Listen to the following speech and identify the emotion expressed by the speaker.\\
            (1) Angry \enspace (2) Disgust \enspace (3) Fear \enspace (4) Joy \enspace (5) Sadness \enspace (6) Surprised
            \vspace{4pt}
        \end{minipage}
        }
    \end{minipage}

    \vspace{6pt}

    \begin{minipage}{0.95\columnwidth}
    \fbox{
        \begin{minipage}{0.93\columnwidth}
            \fontsize{10}{10}\selectfont
            \textbf{Human Annotator Instruction: Global Accent} \\
            \rule{\textwidth}{0.4pt}
            \fontsize{9}{10}\selectfont  
            Speech: \{speech\}\\
            Listen to the following speech and identify the English accent of the speaker.\\
            (1) India \enspace (2) UK \enspace (3) Australia
            \vspace{4pt}
        \end{minipage}
        }
    \end{minipage}

    \vspace{6pt}

    \begin{minipage}{0.95\columnwidth}
    \fbox{
        \begin{minipage}{0.93\columnwidth}
            \fontsize{10}{10}\selectfont
            \textbf{Human Annotator Instruction: Timbre} \\
            \rule{\textwidth}{0.4pt}
            \fontsize{9}{10}\selectfont  
            Speech: \{speech\}\\
            Listen to the following speech and identify the gender and age group of the speaker.\\
            (1) Adult Female \enspace (2) Adult Male \enspace (3) Elderly Female \enspace (4) Elderly Male
            \vspace{4pt}
        \end{minipage}
        }
    \end{minipage}
    \caption{Human annotator instructions for abstract acoustic feature validation used in Section \ref{subsec:task-decomp}.}
    \label{app-fig:human-instruction}
\end{figure}

\begin{figure}[t]  
\centering
\includegraphics[width=1.0\columnwidth, keepaspectratio]{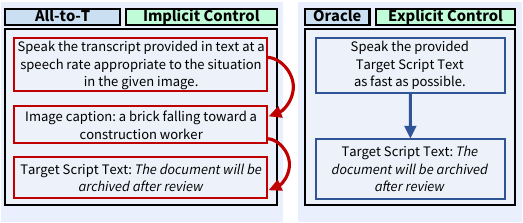}
\caption{
Illustration of the All-to-T and Oracle conditions used in Controlled Input Decomposition.
All-to-T textualizes all inputs while still requiring inference of the target acoustic value from context, whereas Oracle makes the target value explicit in the instruction.
}
\label{fig:all-to-T-vs-Oracle}
\end{figure}

\begin{figure}[t]  
\centering
\includegraphics[width=1.0\columnwidth, keepaspectratio]{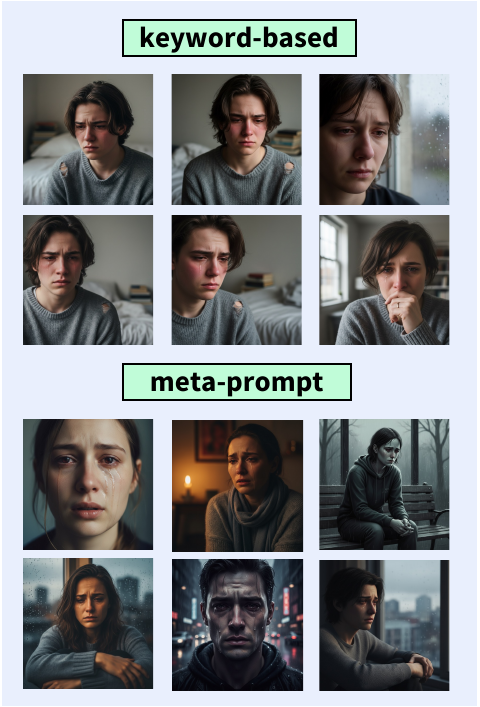}
\caption{
Image diversity comparison between keyword-based and meta-prompt generation for the keyword ``a person with a sad expression.'' 
All images in this example are generated with Gemini 2.5 Flash Image.
}
\label{fig:keyword-metaprompt}
\end{figure}

\begin{figure*}[t]  
\centering
\includegraphics[width=1\textwidth, keepaspectratio]{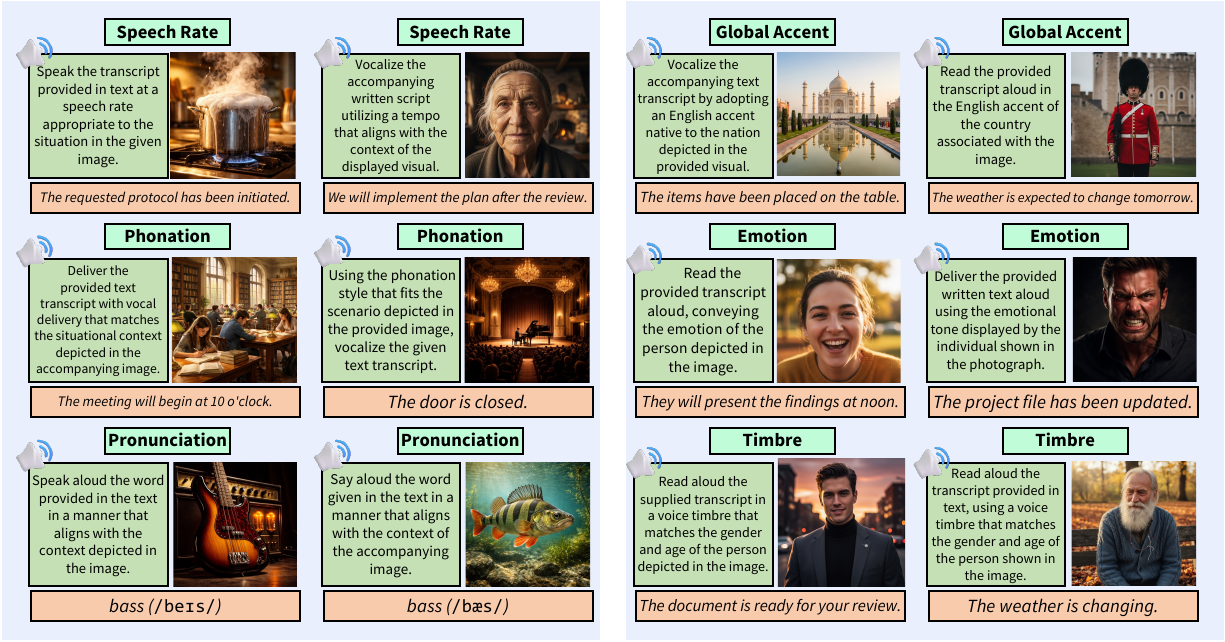}
  \caption{
Examples of \benchmark\ data for each acoustic feature.
}
\label{fig:app-data-examples}
\end{figure*}

\begin{figure*}[t]  
\centering
\includegraphics[width=1\textwidth, keepaspectratio]{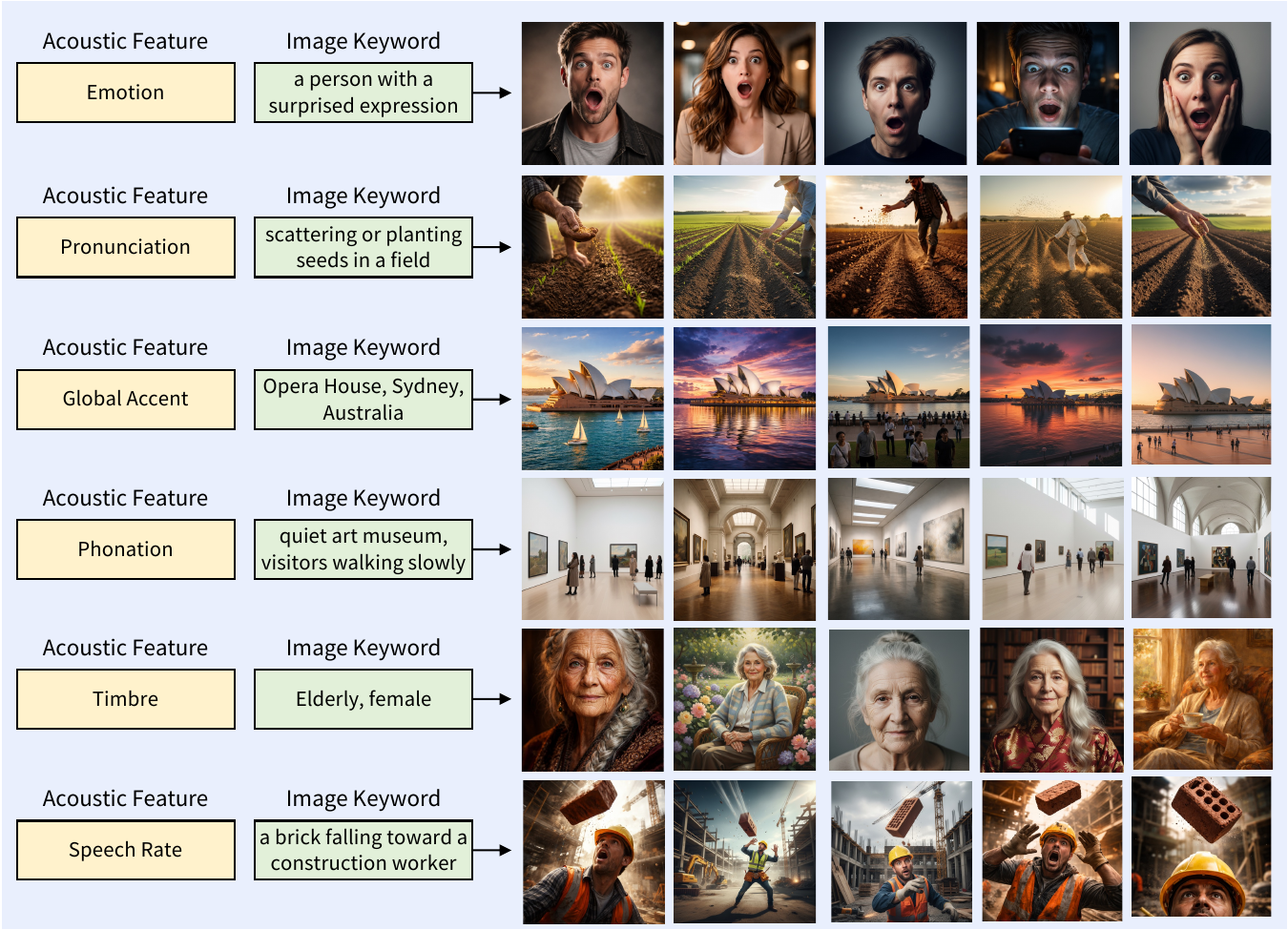}
  \caption{
Examples of image keywords and representative final generated images for each image keyword.
}
\label{fig:app-aco-features-keyword-images}
\end{figure*}

\begin{figure*}[t]  
\centering
\includegraphics[width=1\textwidth, keepaspectratio]{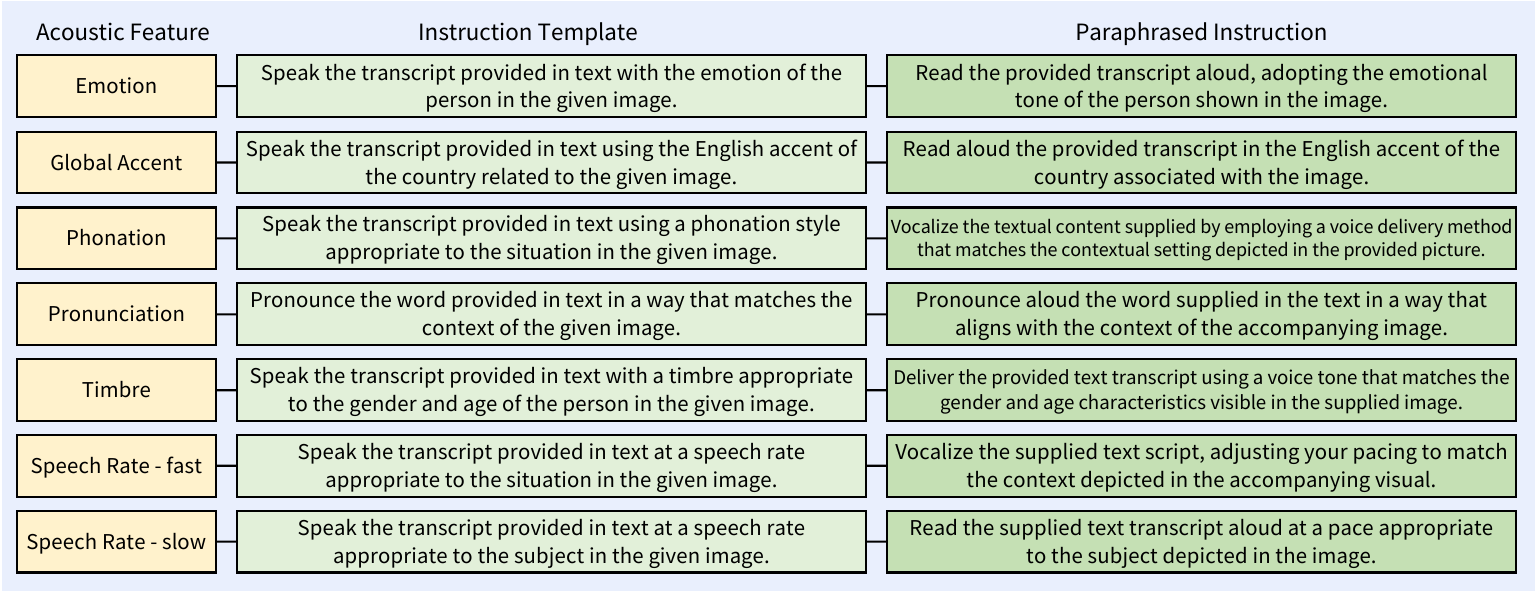}
  \caption{
Example paraphrases of spoken instruction templates.
}
\label{fig:app-inst-paraphrase-examples}
\end{figure*}

\begin{figure*}[t]  
\centering
\includegraphics[width=1\textwidth, keepaspectratio]{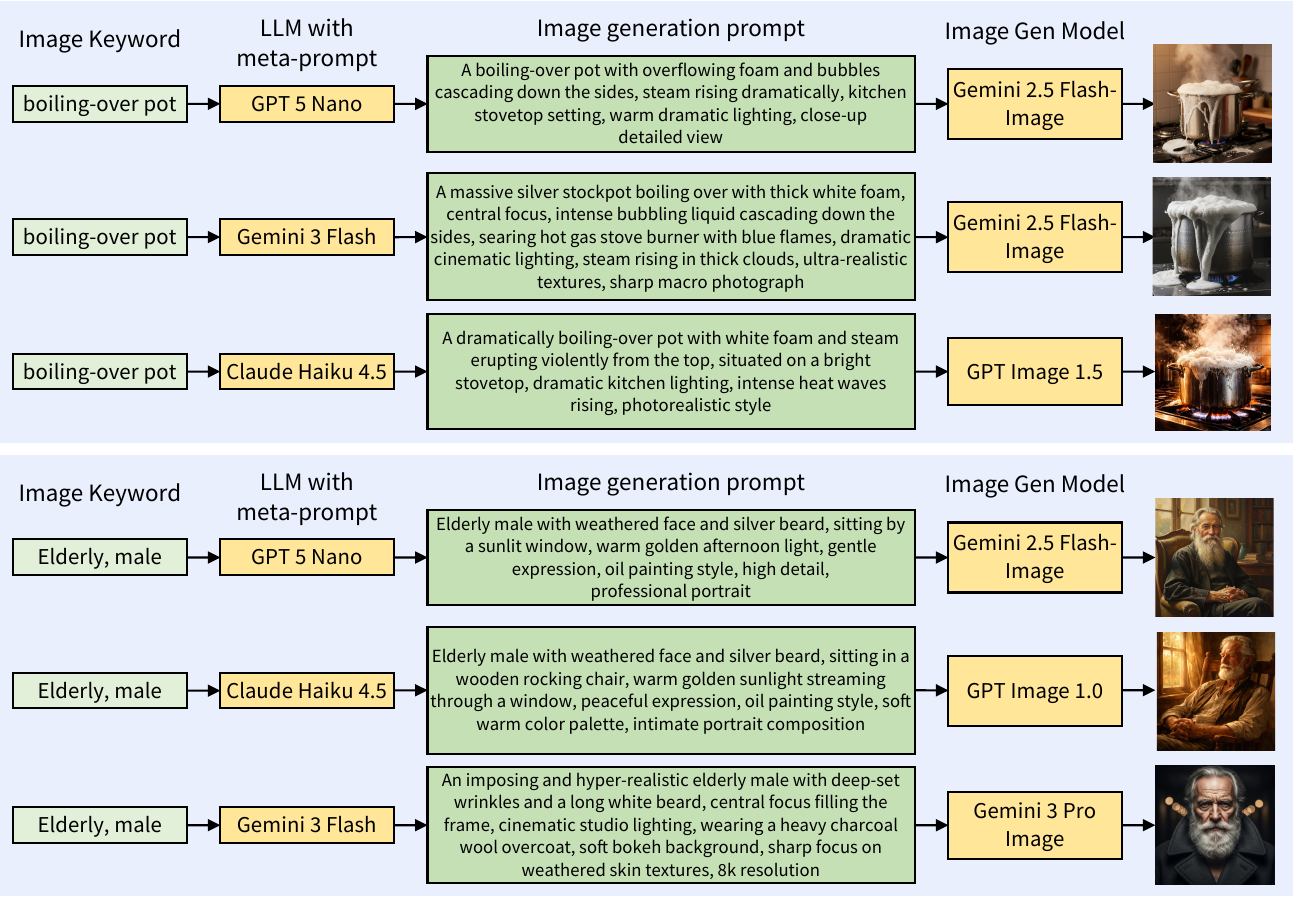}
  \caption{
Examples of meta-prompt-based image prompt expansion.
}
\label{fig:app-keyword-meta-images}
\end{figure*}

\begin{figure*}[t]  
\centering
\includegraphics[width=1.0\textwidth, keepaspectratio]{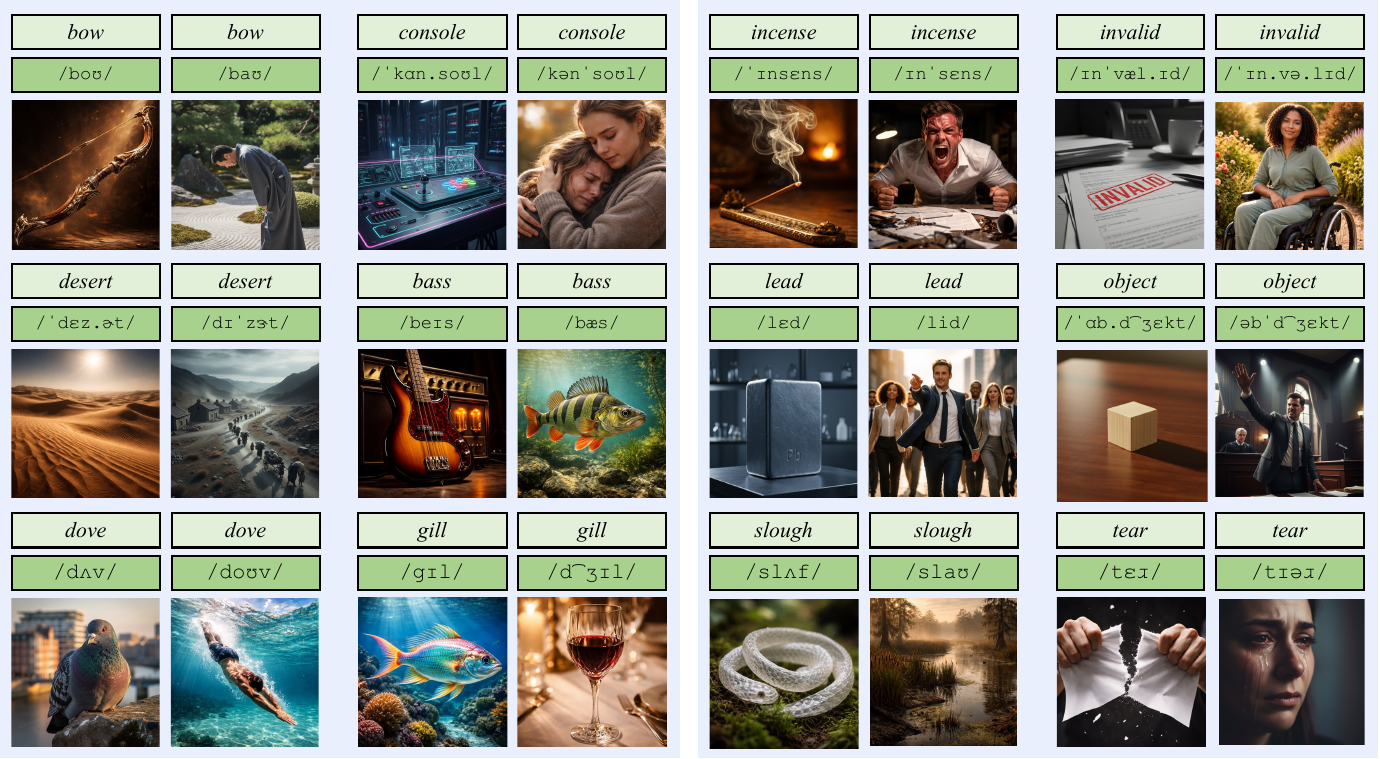}
  \caption{
Image examples of pronunciation features.
}
\label{fig:app-pronunciation-examples}
\end{figure*}

\end{document}